\documentclass{article}

\usepackage[main, preprint]{neurips_2026}

\usepackage[utf8]{inputenc} 
\usepackage[T1]{fontenc}    
\usepackage{hyperref}       
\usepackage{url}            
\usepackage{booktabs}       
\usepackage{amsfonts}       
\usepackage{nicefrac}       
\usepackage{microtype}      
\usepackage{xcolor}         
\usepackage{graphicx}
\usepackage{pifont}
\usepackage{amsmath}
\usepackage{amssymb}
\usepackage{marvosym}
\usepackage{appendix}
\usepackage{subcaption}
\usepackage{enumitem}
\usepackage{array}
\usepackage{tcolorbox}
\tcbuselibrary{listings}

\title{Beyond the Cartesian Illusion: Testing Two-Stage Multi-Modal Theory of Mind under Perceptual Bottlenecks}

\author{%
  Yajing Zhou \textsuperscript{$\dagger$}, Xiangyu Kong\textsuperscript{$\dagger$\Letter} \\
  College of Computer Science, Beijing Information Science and Technology University\\
  No.55 Taihang Road, Changping District, Beijing, 102206, China\\
  \textsuperscript{$\dagger$} Equally contributed \Letter Corresponding to \texttt{xykong@bistu.edu.cn}\\
}

\begin{document}

\maketitle

\begin{abstract}
  While Multi-Modal Large Language Models (MLLMs) demonstrate impressive capabilities in general reasoning, their embodied spatial intelligence remains hampered by a "Cartesian Illusion"—a reliance on text-based probability distributions that lack grounded, 3D topological understanding. This limitation is starkly exposed in multi-agent environments, which demand more than just scene perception; they require second-order Theory of Mind (ToM). Specifically, an Agent A must be able to infer Agent B's belief about the environment, governed strictly by Agent B's physical orientation and sensory limitations. In this paper, we probe the limits of two-stage spatial inference in MLLMs through a novel audio-visual task: requiring Agent A to predict Agent B’s estimation of A’s relative location. To solve this, we propose an Epistemic Sensory Bottleneck module that abandons rigid, rule-based coordinate transformations. Instead, we introduce an Anchor-Based Embodied Spatial Decomposition Chain-of-Thought (CoT). This guides the MLLM through a "geometric-to-semantic" projection, forcing it to first establish B's local coordinate system and then dynamically weight visual and auditory modalities based on whether A falls within B's visual frustum. Extensive evaluations reveal that while current MLLMs fundamentally struggle with spatial symmetry and out-of-view ambiguities (establishing a rigorous zero-shot baseline of 42\% accuracy), our sensory-bounded reasoning chain robustly outperforms pure egocentric and allocentric baselines. By systematically benchmarking these perceptual bottlenecks, our work exposes the current limits of MLLM spatial reasoning and establishes a foundational paradigm for epistemic, modality-aware inference in Embodied AI.
\end{abstract}

\begin{figure}[t]
  \centering
  \includegraphics[width=\linewidth]{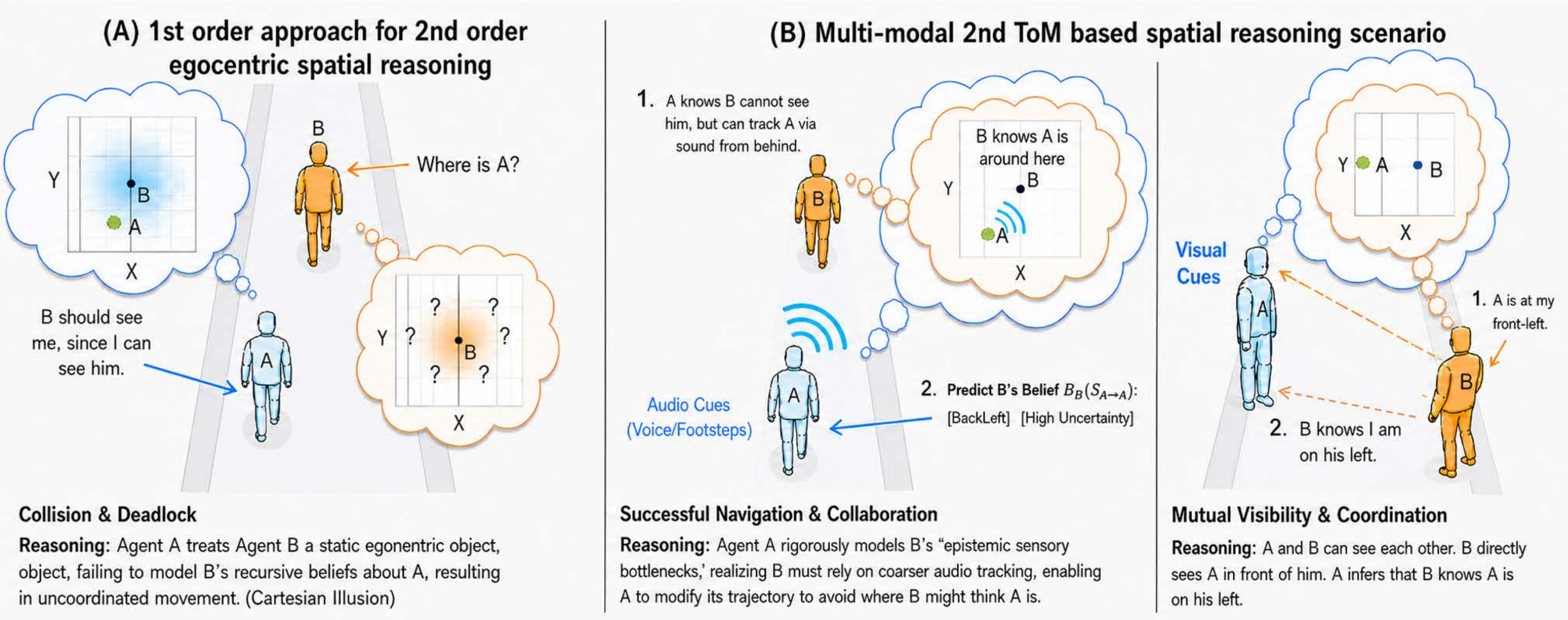}
  \caption{Comparison of the Cartesian Illusion Dilemma vs. the Proposed Embodied 2nd Order ToM Reasoning Task. (A) Traditional MLLMs succumb to the Cartesian Illusion, assuming an omniscient global coordinate system where spatial relations are perfectly symmetrical. (B) In our proposed task, Agent A must predict Agent B's internal belief state $\left(Belief_B\left(Pos_A\right)\right)$. Because Agent A falls outside Agent B's visual frustum (FOV), the inference requires a Perspective Shift $\left(PI_A\rightarrow B\right)$ that explicitly models sensory bottlenecks, forcing Agent B (and thus Agent A's simulation of B) to rely on Spatial Audio and Self-Motion rather than visual confirmation.}
  \label{fig:background}
\end{figure}

\section{Introduction}
\label{sec:intro}

The "Cartesian Illusion"—the implicit assumption that cognitive agents possess perfect, omnidirectional access to a shared, mathematically symmetrical global coordinate system—has long plagued the development of embodied artificial intelligence \cite{dreyfus1992what, brooks1991intelligence}. While this abstraction simplifies algorithmic design, it fundamentally violates the physical realities of embodiment, where perception is strictly bounded by field of view, occlusion, and sensory degradation \cite{gibson2014ecological}. Overcoming this illusion is not merely a theoretical exercise; it is a critical, long-standing bottleneck whose resolution bears immense generalizability and application potential for achieving robust, human-like multi-agent collaboration in complex physical environments \cite{linsley20243d, puig2020watch}.

This illusion becomes profoundly fatal in the specific context of Recursive (Second-Order) Theory of Mind (ToM). In this challenging setting, an observer (Agent A) must predict the internal belief state of a target (Agent B) regarding Agent A’s own spatial location, relying entirely on Agent A's partial, egocentric multi-modal sensor stream. To succeed, Agent A cannot simply "locate" Agent B in a global grid; Agent A must rigorously simulate how the world appears—and sounds—from Agent B's heavily constrained vantage point \cite{flavell2013perspectives}. This requires Agent A to model not just the geometry of the scene, but the epistemic sensory bottleneck of the other.

This "Self-in-Other's-Eye" reasoning is a fundamental hallmark of social intelligence \cite{premack1978does, baron1985does}, yet it is notoriously difficult for current models. Consider a common "hallway dance": when two people approach each other, their coordination depends not on a shared GPS coordinate, but on each person predicting where the other thinks they are \cite{kruse2013human}. For an AI to be truly "savvy," it must move beyond first-order perception (e.g., "Where is the object?") to a recursive understanding of perceptual fallibility \cite{apperly2009humans} (e.g., "Where does B think I am, given B is currently facing away and can only hear my footsteps?"). In this work, we move beyond the Cartesian Illusion by proposing a two-stage multi-modal ToM framework that explicitly models these sensory constraints, marking a significant step toward agents that reason with a human-like awareness of each other’s internal states.

Previous foundational works have made remarkable strides in Embodied Question Answering (EQA) \cite{das2018embodied, shridhar2020alfred, majumdar2024openeqa} and text-based ToM \cite{mao2024review, sap2022neural, ullman2023large}. State-of-the-art vision-language and audio-language models excel at projecting multi-modal tokens into 3D spaces \cite{grauman2022ego4d, liu2023visual, Chen_2024_CVPR} and utilizing spatial audio for first-order target tracking \cite{chen2025savvy, gao2020listen}. However, when subjected to this second-order spatial reasoning scenario, these otherwise highly capable models collapse. Because they process environmental inputs end-to-end without explicit cognitive modality constraints, they inherently succumb to the Cartesian Illusion. They implicitly assume Agent B shares Agent A's geometric awareness, leading to catastrophic ``spatial flip curses'' (e.g., confusing left and right during perspective shifts) and causing models to hallucinate that Agent B can visually localize Agent A even when A is entirely outside of B's visual horizon.

Based on the simple but profound intuition that an agent's internal beliefs are strictly bounded by its \textit{epistemic sensory bottleneck}, we propose a solution that explicitly shatters the Cartesian Illusion. We introduce the \textbf{Observe-to-Believe Pipeline}, an explicit multi-stage framework for recursive ToM reasoning in embodied systems. Rather than forcing a single network to implicitly untangle spatial coordinate transforms and sensory degradation simultaneously \cite{bubeck2023sparks}, we explicitly disentangle geometric observation from epistemic inference. 

\begin{figure}[t]
  \centering
  \includegraphics[width=\linewidth]{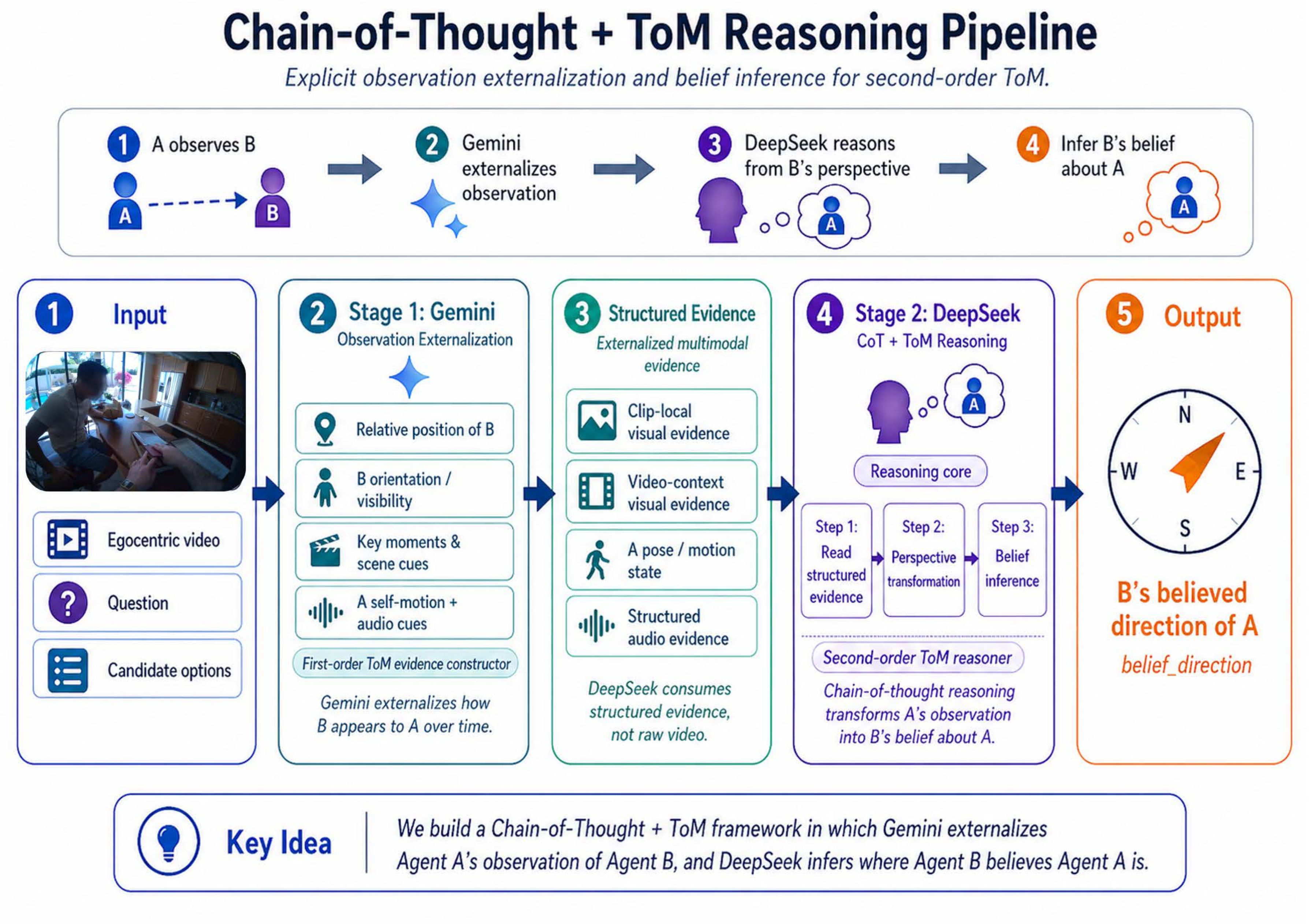}
  \caption{The Observe-to-Believe Pipeline shatters the Cartesian Illusion. Stage I extracts structured visual evidence from Agent A's egocentric stream. Stage II performs a modality-aware perspective shift via spatial horizon conversion, dynamically routing inference through visual or audio pathways based on Agent B's inferred epistemic sensory bottleneck.}
  \label{fig:pipeline}
\end{figure}

As illustrated in Figure~\ref{fig:pipeline}, the pipeline operates via a dual-stage architecture. In Stage I (ToM-Oriented Observation Modeling), a Vision-Language Model extracts structured physical evidence from Agent A's egocentric stream, determining Agent B's visibility, relative orientation, and spatial stability. In Stage II (Belief-Oriented ToM Inference), a Large Language Model executes a rigorous perspective shift ($\pi_{A \rightarrow B}$). Crucially, this shift is governed by an explicit \textbf{spatial horizon conversion}. If B's visual horizon fails to capture A, the framework dynamically shifts its reasoning pathway, simulating B's reliance on fused spatial audio and self-motion evidence rather than visual priors. 

Through rigorous evaluation, we demonstrate that while standard egocentric and allocentric baselines perform adequately when agents share a line-of-sight, they fail drastically in degraded conditions. Our pipeline excels precisely in these challenging, invisible scenarios, clearly validating our intuition.

In summary, our main contributions are:
\begin{itemize}
    \item \textbf{Addressing the Cartesian Illusion:} We identify and formulate the Cartesian Illusion in embodied ToM, proposing the novel, training-free \textit{Observe-to-Believe} architecture that explicitly separates physical feature extraction from recursive belief inference.
    \item \textbf{Sensory-Bounded Belief Modeling:} We introduce a dynamic reasoning policy that successfully models epistemic sensory bottlenecks. By executing an explicit spatial horizon conversion, the framework automatically shifts between audio-visual and audio-only reasoning branches based on visual frustum constraints.
    \item \textbf{Superiority in Challenging and Occluded Scenarios:} Extensive experiments demonstrate that our two-stage ToM reasoning significantly outperforms traditional baselines in challenging ``invisible'' cases, empirically proving the absolute necessity of modality-aware spatial horizon conversion for accurate multi-agent belief modeling.
\end{itemize}

\section{Related Work}
\label{sec:related_work}

Our research sits at the intersection of embodied spatial reasoning, audio-visual learning, and cognitive Theory of Mind (ToM). In this section, we review recent advancements across these domains, highlighting the transition from first-order, single-agent perception to the necessity of second-order, multi-modal belief modeling.

\subsection{Embodied Question Answering and Spatial Reasoning}
The past three years have witnessed a burst of research in Embodied Question Answering (EQA) and spatial reasoning, driven by the integration of Large Language Models (LLMs) and Vision-Language Models (VLMs) with 3D environments \cite{driess2023palm, zitkovich2023rt}. Foundational works like 3D-LLM \cite{hong20233dllm} and LEO \cite{huang2023embodied} pioneered the injection of 3D point clouds into language models, while more recent models such as Video-3D LLM \cite{zheng2025video} treat scenes as dynamic videos to align representations with real-world spatial contexts. Contemporary methods have evolved from navigating static, fully-observable grids \cite{anderson2018vision} to performing complex spatial grounding in dynamic, egocentric datasets like Ego4D \cite{grauman2022ego4d}, EPIC-Kitchens \cite{damen2022rescaling}, and SQA3D \cite{masqa3d}. 

\textbf{Limitations:} Despite these advances, the vast majority of current EQA tasks and environments \cite{majumdar2024openeqa, Chen_2024_CVPR} remain fundamentally \textit{first-order} and \textit{single-agent}. They ask ``Where is the object relative to me?'' or ``Where am I in the room?''. Even when multi-agent scenarios are introduced, models typically rely on a global allocentric map that assumes shared, omnidirectional perception. Our work introduces a novel task that challenges this paradigm, requiring models to perform recursive coordinate transformations based strictly on partial, egocentric observations.

\subsection{Audio-Visual Learning in Embodied Environments}
Spatial audio is a critical modality in embodied AI, particularly for tracking targets outside the visual field of view (FoV). Platforms like SoundSpaces \cite{chen2020soundspaces} and its successors \cite{gan2020look} catalyzed a wave of audio-visual navigation tasks. Concurrently, audio-visual QA benchmarks \cite{gao2020listen, yun2021pano} have driven the development of multi-modal fusion techniques like AST \cite{gong2021ast}. Recently, large audio language models such as SALMONN \cite{tang2023salmonn} and Qwen-Audio \cite{chu2023qwen} have demonstrated profound abilities to fuse auditory signals with linguistic reasoning. Most directly related to our work is the SAVVY framework \cite{chen2025savvy}, a recent state-of-the-art benchmark for first-order egocentric spatial track estimation that seamlessly correlates spatial audio cues with visual bounding boxes to resolve dynamic queries.

\textbf{Limitations:} Existing audio-visual frameworks treat audio primarily as a compensatory signal for the \textit{self} (e.g., Agent A uses audio to find an occluded object). They do not treat audio as an \textit{epistemic constraint for others}. Our framework pioneers the use of audio-visual signals to model a ``sensory bottleneck.'' We transition spatial audio from a mere tracking feature into a cognitive gating mechanism, determining when Agent B's internal belief state must degrade from high-resolution visual geometry to low-resolution acoustic estimation.

\subsection{Theory of Mind and Visual Perspective Taking}
Theory of Mind in AI has historically been evaluated using text-based narratives (e.g., ToMi \cite{le2019revisiting}, SocialIQA \cite{sap2019socialiqa}) or simplified 2D grid-worlds \cite{rabinowitz2018machine}. Despite the impressive deductive capabilities of modern LLMs \cite{bubeck2023sparks}, recent studies show they routinely fail at trivial ToM tasks when physical constraints are introduced \cite{ullman2023large, sclar2023quantifying}. Recently, the community has pushed ToM into continuous 3D environments to evaluate Visual Perspective Taking (VPT), a concept deeply rooted in developmental psychology \cite{flavell2013perspectives}. Benchmarks such as the 3D Perception Challenge (3D-PC) \cite{linsley20243d} have demonstrated that while modern DNNs can understand depth and object order, they fail drastically when required to perform Level-1 and Level-2 VPT \cite{labash2020perspective}—determining what another agent can see and how objects appear from their vantage point. 

\textbf{Limitations:} Current ToM methodologies suffer from two critical flaws when applied to embodied agents. First, end-to-end LLMs inherently struggle with spatial geometry, often falling victim to the ``spatial flip curse'' during perspective shifts. Second, current second-order ToM models implicitly assume the target agent possesses perfect information, failing to account for the physical limitations of embodiment. Our proposed \textit{Observe-to-Believe} pipeline bridges this gap. By utilizing a two-stage architecture that explicitly calculates the spatial horizon before performing ToM inference, we introduce the first framework capable of \textit{Sensory-Bounded 2nd-Order ToM}, accurately predicting the systematic localization errors agents make in noisy or occluded environments.

\section{Methodology}
\label{sec:method}

To tackle the challenge of second-order spatial perspective taking, we propose the \textbf{Observe-to-Believe Pipeline}, a two-stage framework that explicitly disentangles geometric observation from modality-aware belief inference.

\subsection{Problem Formulation: Recursive Spatial Theory of Mind}
\label{subsec:problem_formulation}

Consider two embodied agents, an observer $A$ and a target $B$, operating in a shared 3D environment. Let $P_A(t)$ and $P_B(t)$ denote their true poses in world coordinates at time $t$. In a first-order spatial reasoning task, Agent $A$ utilizes its egocentric sensor stream $\mathcal{O}_A = \{V_A, A_A, M_A\}$ (comprising visual frames, spatial audio, and ego-motion/IMU) to estimate $B$'s relative state $S_{A \rightarrow B}$. 

Our task, however, extends to \textbf{Recursive (Second-Order) Theory of Mind}. Agent $A$ must predict \textit{Agent $B$'s internal belief} regarding $A$'s relative location, denoted as $\mathcal{B}_{B}(S_{B \rightarrow A})$, relying strictly on $A$'s egocentric observations $\mathcal{O}_A$. 

The core mathematical challenge lies in the lack of direct access to $B$'s sensor stream $\mathcal{O}_B$. Therefore, $A$ must construct a simulated observation $\hat{\mathcal{O}}_B$ by applying a perspective shift $\pi_{A \rightarrow B}$. Crucially, this shift is not a simple coordinate inversion; it is strictly bounded by $B$'s \textbf{epistemic sensory bottleneck}. We formulate $B$'s belief state as a conditional probability distribution:
\begin{equation}
    \mathcal{B}_{B}(S_{B \rightarrow A}) = \arg\max_{S} P\left(S \mid \pi_{A \rightarrow B}(\mathcal{O}_A), \Phi_B \right)
\end{equation}
where $\Phi_B$ represents $B$'s perceptual constraints (e.g., visual field of view). To realize this, our pipeline decomposes the inference into two explicit stages: ToM-Oriented Observation Modeling (Stage I) and Belief-Oriented ToM Inference (Stage II).

\subsection{Stage I: ToM-Oriented Observation Modeling}
\label{subsec:stage1}

In Stage I, Agent $A$ extracts structured, physically grounded evidence from its egocentric stream. Building upon the foundational audio-visual tracking capabilities of SAVVY, we employ a Vision-Language Model (VLM) as an observation engine to externalize socially relevant visual evidence.

Given the visual frame sequence $V_A$, the VLM identifies Agent $B$ and extracts the relative coordinate vector $p_{B|A}$ and, critically, $B$'s facing direction in $A$'s local coordinate frame, denoted as $\hat{\theta}_B$. Instead of directly predicting spatial labels, Stage I outputs a structured evidence tuple $E_{vis} = \langle I_{B,vis}, \theta_{B \rightarrow A}, \Delta_{pos} \rangle$:
\begin{itemize}
    \item $\theta_{B \rightarrow A}$: The geometric orientation of $B$ relative to $A$. If $A$'s position is the origin $(0,0)$, we estimate the relative observation angle $\alpha_{A|B}$, which defines $A$'s position relative to $B$'s forward heading vector.
    \item $I_{B,vis}$: A boolean visibility state indicating whether $A$ is currently looking at $B$'s front/profile or $B$'s back.
    \item $\Delta_{pos}$: Contextual stability markers, tracking spatial memory derived from $A$'s ego-motion ($a_{world}, a_{orient}$).
\end{itemize}

Simultaneously, spatial audio evidence ($A_A$)—such as Interaural Time/Level Differences (ITD/ILD) from footsteps or speech—is extracted to form an auxiliary audio cue, ensuring tracking continuity even when visual evidence degrades.

\subsection{Stage II: Belief-Oriented ToM Inference via Spatial Horizon Conversion}
\label{subsec:stage2}

The core novelty of our framework resides in Stage II, where a Large Language Model (LLM) is prompted to act as a cognitive reasoning engine. Rather than implicitly mapping $A$'s view to $B$'s belief, Stage II enforces an explicit \textbf{Spatial Horizon Conversion} guided by a modality-aware reasoning policy.

To model $B$'s epistemic state, we define $B$'s visual Field of View (FoV) as a frustum parameterized by angle $\Phi$ (e.g., $120^\circ$). We introduce a binary sensory mask $M_v$ (the visibility condition), which Agent $A$ computes geometrically from Stage I outputs:
\begin{equation}
    M_v = \mathbb{I} \left( |\alpha_{A|B}| \le \frac{\Phi}{2} \right)
\end{equation}
This mask dictates the reasoning pathway the LLM must follow to infer $\mathcal{B}_{B}$:

\paragraph{Pathway 1: Audio-Visual Fusion (In-View Condition, $M_v = 1$)}
When $A$ determines it falls within $B$'s visual horizon, $A$ infers that $B$ possesses high-fidelity, cross-modal perception. The LLM executes a direct coordinate perspective shift based on the structured orientation cues:
\begin{equation}
    \hat{S}_{B \rightarrow A} \approx R(-\hat{\theta}_B) \cdot (-p_{B|A})
\end{equation}
where $R$ is the 2D rotation matrix. The LLM's prompt policy dictates high confidence, predicting precise spatial labels (e.g., ``Front-Right'').

\paragraph{Pathway 2: Sensory Bottleneck via Audio/Motion (Out-of-View Condition, $M_v = 0$)}
The critical failure point of traditional baselines occurs when $M_v = 0$ (e.g., $A$ is behind $B$). Here, the visual horizon fails. Our reasoning policy forces the LLM to drop visual priors for $B$ and rely entirely on fused audio evidence and $A$'s ego-motion compensation:
\begin{equation}
    \mathcal{B}_B(S_{B \rightarrow A}) = \text{LLM\_Reasoning}\left( \text{Audio}(A_A), \Delta_{pos} \right)
\end{equation}
In this branch, the LLM simulates $B$'s reliance on spatial audio localization. By explicitly acknowledging this sensory degradation, the framework naturally models the inherent ambiguities of non-visual localization (e.g., front-back confusion or left-right acoustic shadowing). 

By formulating the second-order ToM task as an explicit, conditionally gated equation based on the sensory horizon:
\begin{equation}
    \mathcal{B}_B = M_v \cdot \mathcal{F}_{visual\_dominant}(E_{vis}) + (1 - M_v) \cdot \mathcal{F}_{audio\_dominant}(A_A, \Delta_{pos})
\end{equation}
our Observe-to-Believe pipeline forces the reasoning model to respect the physical constraints of embodiment. This mechanism fundamentally prevents the omnidirectional hallucinations common in standard LLMs, yielding robust predictions even in highly noisy or occluded multi-agent interactions.

\section{Experiment}
\label{sec:experiment}

To evaluate the effectiveness of our proposed \textbf{Observe-to-Believe} pipeline, we conduct extensive experiments focusing on the complex spatial reasoning task of predicting Agent B's belief of Agent A's location based on Agent A's egocentric observations. All ground-truth annotations and evaluation metrics are derived from the SAVVY dataset\cite{chen2025savvy}. All experiments are conducted with a 16GB RTX-4090 laptop machine. 

\subsection{Experimental Setup}
\textbf{Datasets and Tasks.} We utilize the \textit{ego\_direction} subset of the SAVVY dataset as our primary evaluation testbed, where Agent A attempts to infer B's perspective. To test out-of-distribution robustness, we also evaluate a baseline on the \textit{exo\_direction} subset. The target task requires the model to predict the relative spatial orientation. While the tasks inherently contain both simple (3-way) and hard (4-way) configurations, we report the overall accuracy to measure general ToM reasoning capabilities. Based on the two versions of first order reasoning methods in SAVVY\cite{chen2025savvy} respectively, We define two end-to-end reasoning baselines where a Vision-Language Model (Gemini) is directly prompted to output the final ToM prediction:

\quad \textbf{Baseline 1 (SAVVY Egocentric):} Directly infers B's belief of A with \textit{ego\_direction} data.

\quad \textbf{Baseline 2 (SAVVY Allocentric):} Directly infers B's belief of A with \textit{exo\_direction} data.

\textbf{Implementation Details of Our Pipeline.} To explicitly demonstrate the modularity and decoupled nature of our Observe-to-Believe framework, we employ a heterogeneous model architecture. For Stage I (ToM-Oriented Observation Modeling), we deploy \textbf{Gemini-2.5-Pro} to act as the multi-modal sensory cortex, extracting structured physical and geometric evidence directly from the raw egocentric video stream. For Stage II (Belief-Oriented ToM Inference), we route this structured evidence into \textbf{DeepSeek-v4-Flash}, leveraging its robust logical reasoning capabilities to execute the modality-aware perspective shifts. This hybrid setup not only ensures optimal performance at both the perceptual and cognitive levels but also validates that our explicit spatial horizon conversion can be seamlessly generalized across different foundation models. To thoroughly investigate the impact of information granularity, we design varying contextual conditions for Stage II: \textit{Partial Context} (retaining only core states: \texttt{is\_static}, \texttt{visibility\_to\_camera}, \texttt{b\_orientation}, and A's pose) and \textit{Full Context} (adding rich spatial anchors such as \texttt{distance}, \texttt{direction}, and environmental landmark descriptions).

\begin{figure}[t]
  \centering
  
  \begin{minipage}[b]{0.47\textwidth}
    \centering
    \small 
    \setlength{\tabcolsep}{6pt} 
    \renewcommand{\arraystretch}{1.25}
    \begin{tabular}{lccc}
    \toprule
    \textbf{Method} & \textbf{Context} & \textbf{Audio} & \textbf{Acc.} \\ \midrule
    Baseline 2 (\textit{Exo}) & None & \ding{51} & 0.2442 \\
    Baseline 1 (\textit{Ego}) & None & \ding{51} & 0.3436 \\ \midrule
    Ours (Exp 3) & Partial & \ding{55} & 0.4978 \\
    Ours (Exp 2) & Partial & \ding{51} & 0.4846 \\
    Ours (Exp 5) & Full & \ding{51} & 0.5000 \\
    \textbf{Ours (Exp 4)} & \textbf{Full} & \ding{55} & \textbf{0.5066} \\ \bottomrule
    \end{tabular}
    
    \captionof{table}{Overall performance and ablation study on Stage II contextual richness. Our disentangled pipeline achieves peak accuracy (0.5066) by leveraging full geometric context.}
    \label{tab:overall_performance}
  \end{minipage}
  \hfill 
  \begin{minipage}[b]{0.45\textwidth}
    \centering
    \includegraphics[width=1.0\linewidth]{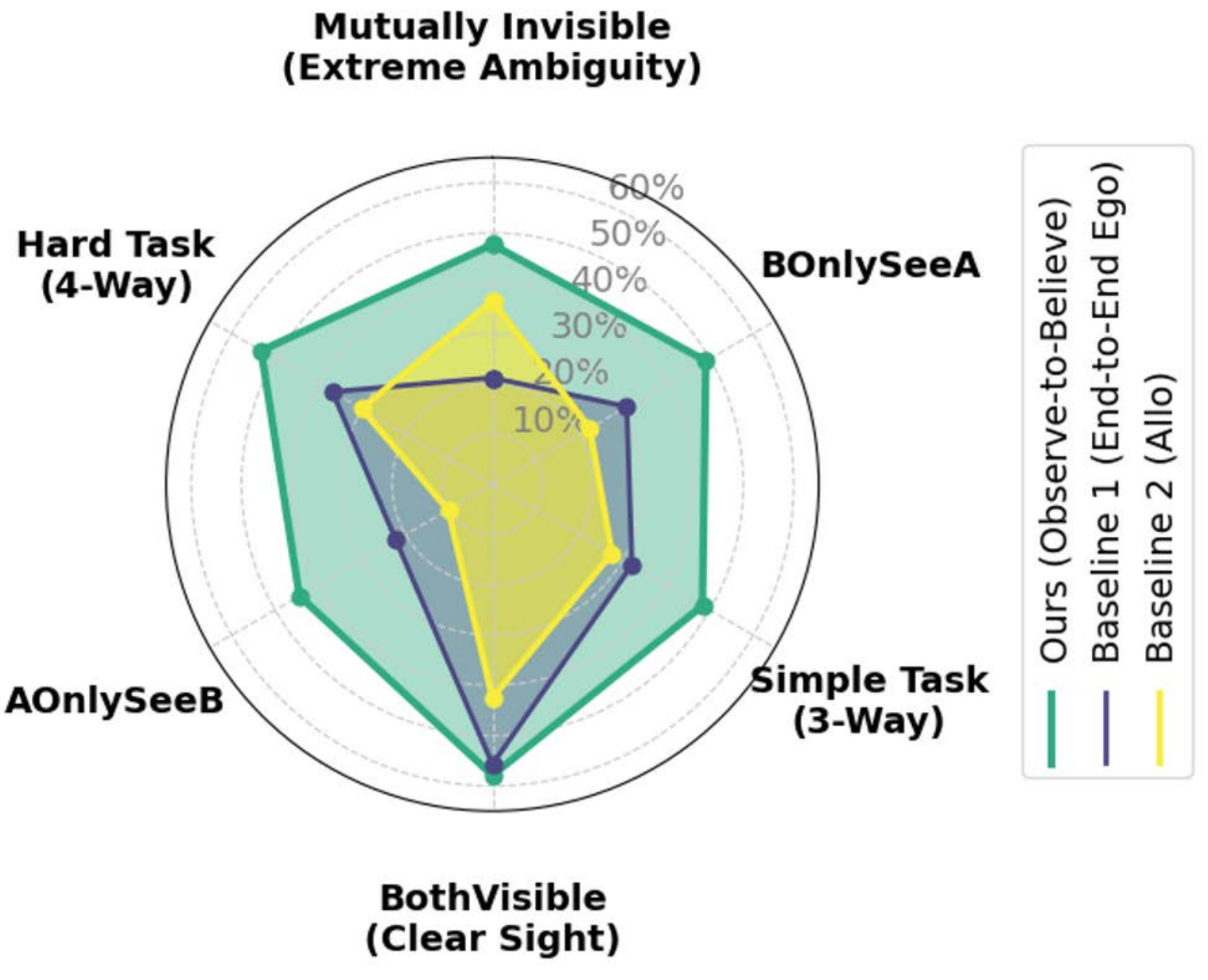} 
    \caption{We compare our Observe-to-Believe pipeline against Baseline 1 (Egocentric) and Baseline 2 (Exocentric).}
    \label{fig:hexagon_radar}
  \end{minipage}
  
\end{figure}

\subsection{Overall Performance Evaluation}
In Stage I, the VLM is tasked with extracting structural physical evidence, i.e. Agent B's relative orientation (\texttt{b\_orientation\_to\_camera}) from the video. We evaluated this stage under two modalities: video-only and video-with-audio. The model achieved an accuracy of \textbf{0.5966} using video-only inputs, while the addition of audio slightly decreased performance to \textbf{0.5088}. This performance gap suggests potential modality alignment noise when directly processing raw audio-visual streams within the VLM, underscoring the necessity of our explicit routing strategy in Stage II, which treats extracted spatial clues rather than raw multi-modal inputs as the basis for higher-order ToM reasoning.
Table \ref{tab:overall_performance} presents the quantitative results. The end-to-end Baseline 1 yields a low accuracy of 0.3436, indicating that implicitly learning perspective shifting and modality degradation simultaneously is insufficient for recursive ToM. By disentangling the observation from inference, our pipeline significantly outperforms the baseline. Even under the \textit{Partial Context} setting without audio cues, accuracy surges to 0.4978. When scaling to \textit{Full Context} (introducing explicit 3D spatial anchors and landmark descriptions), the pipeline achieves its peak accuracy of \textbf{0.5066}. This substantial leap (+16.3\% absolute improvement over Baseline 1) empirically validates our core hypothesis: higher-order spatial reasoning requires explicitly structured geometric foundations before applying perspective shifts.

\subsection{Multi-Dimensional Performance Analysis}

To rigorously dissect model capabilities, we evaluate accuracy across six interconnected dimensions capturing both epistemic constraints (\texttt{Mutually Visible}, \texttt{A Only Sees B (AOnlySeeB)}, \texttt{B Only Sees A (BOnlySeeA)}, \texttt{Mutually Invisible}) and geometric complexity (\texttt{Simple} versus \texttt{Hard} tasks). The resulting hexagonal footprint (Figure \ref{fig:hexagon_radar}) starkly exposes two distinct modes of cognitive failure in traditional end-to-end reasoning. \textbf{Baseline 1 (End-to-End Ego)} suffers from a ``line-of-sight dependency,'' performing adequately under shared visibility (\texttt{Mutually Visible}, 55.7\%) but collapsing entirely under epistemic bottlenecks (bottoming out at 21.2\% in the \texttt{Mutually Invisible} scenario). Conversely, \textbf{Baseline 2 (End-to-End Exo)} reveals a ``perspective-taking deficit.'' Despite possessing an omnidirectional global view, it fails catastrophically at projecting into a specific agent's internal belief state, dropping to a mere 10.2\% in \texttt{AOnlySeeB} and 22.4\% in \texttt{BOnlySeeA} scenarios.
In stark contrast, our explicit \textbf{Observe-to-Believe pipeline} completely encompasses both baselines by disentangling geometric observation from modality-aware cognitive inference. It seamlessly bridges the epistemic gap—more than doubling Baseline 1's accuracy in the extreme \texttt{Mutually Invisible} scenario (47.7\%) and dominating asymmetrical conditions (49.1\% in \texttt{BOnlySeeA}). Furthermore, it exhibits superior handling of geometric complexity (peaking at 53.2\% on \texttt{Hard} tasks), empirically confirming that explicit spatial horizon conversion is strictly necessary for resolving sensory constraints and the spatial flip curse in multi-agent environments. 

\subsection{Ablation Studies \& Inference Time Analysis}
\label{subsec:audio_bottleneck}



While globally integrating audio cues slightly degrades overall accuracy (Table \ref{tab:overall_performance}), Table \ref{tab:audio_ablation} reveals audio's indispensable role under severe epistemic bottlenecks. In \texttt{Mutually Visible} scenarios, visual cues heavily dominate, and forced audio fusion introduces minor alignment noise. Conversely, in hard cases where the visual horizon is broken (\texttt{Mutually Invisible} and \texttt{AOnlySeeB}), spatial audio acts as a crucial lifeline. Our pipeline successfully leverages these auditory cues to anchor spatial predictions and recover relative geometry when visual priors completely collapse. However, the slight performance drop in shared-horizon conditions exposes a current limitation: gracefully coordinating redundant multi-modal cues without cross-modal interference often remains beyond the logical reasoning capacity of current LLMs, strictly necessitating our explicit spatial routing mechanism.


\begin{figure}[t]
  \centering
  
  \begin{minipage}[b]{0.47\textwidth}
    \centering
    \begin{tabular}{lc}
    \toprule
    \textbf{Condition} & \textbf{$\Delta$ Audio Impact} \\ \midrule
    \texttt{Mutually Visible} & - 0.0073 \\
    \texttt{AOnlySeeB} & \textbf{+ 0.0138} \\
    \texttt{Mutually Invisible} & \textbf{+ 0.0076} \\ \bottomrule
    \end{tabular}
    
    \captionof{table}{Fine-Grained Modality Ablation: compare Exp 4 (w/o Audio) and Exp 5 (w/ Audio).}
    \label{tab:audio_ablation}
  \end{minipage}
  \hfill 
  \begin{minipage}[b]{0.45\textwidth}
    \centering
    \small
    \begin{tabular}{lc}
    \toprule
    \textbf{Method} & \textbf{Average Time (s)} \\ \midrule
    Baseline 1 (Ego) & 38.51 \\ \midrule
    Baseline 2 (Exo) & 35.97 \\ \midrule
    Ours (Stage I) & 59.49 \\
    Ours (Stage II) & 0.92 \\
    Ours (Total) & 60.41 \\ \bottomrule
    \end{tabular}

    \captionof{table}{Average Inference Time Analysis.}
    \label{tab:inference_time}
  \end{minipage}
  
\end{figure}





As detailed in Table \ref{tab:inference_time}, our explicit two-stage architecture introduces only a moderate latency overhead compared to the end-to-end baseline (60.41s vs. 38.51s). Crucially, this temporal increase is entirely localized to the heavy multimodal perceptual extraction executed in Stage I (59.49s). In stark contrast, Stage II performs the complex modality routing and perspective-shifting in a negligible 0.92 seconds. This demonstrates that formalizing an explicit, multi-layered ToM reasoning Chain-of-Thought imposes virtually no computational burden. Given the substantial leap in epistemic robustness and the successful resolution of spatial flip curses across occluded environments, this processing trade-off is highly positive, validating the efficiency and necessity of decoupled spatial inference.

\begin{table*}[htbp]
    \centering
    \renewcommand{\arraystretch}{2.0} 
    \begin{tabular}{ 
        >{\centering\arraybackslash}m{0.09\textwidth} 
        >{\centering\arraybackslash}m{0.3\textwidth} 
        >{\centering\arraybackslash}m{0.3\textwidth} 
        m{0.22\textwidth} 
    }
    \toprule
    \textbf{Setting} & \textbf{Agent A's View} & \textbf{Agent B's View} & \textbf{Analysis \& Results} \\
    \midrule
    
    \textit{A Only Sees B} & 
    \includegraphics[width=\linewidth]{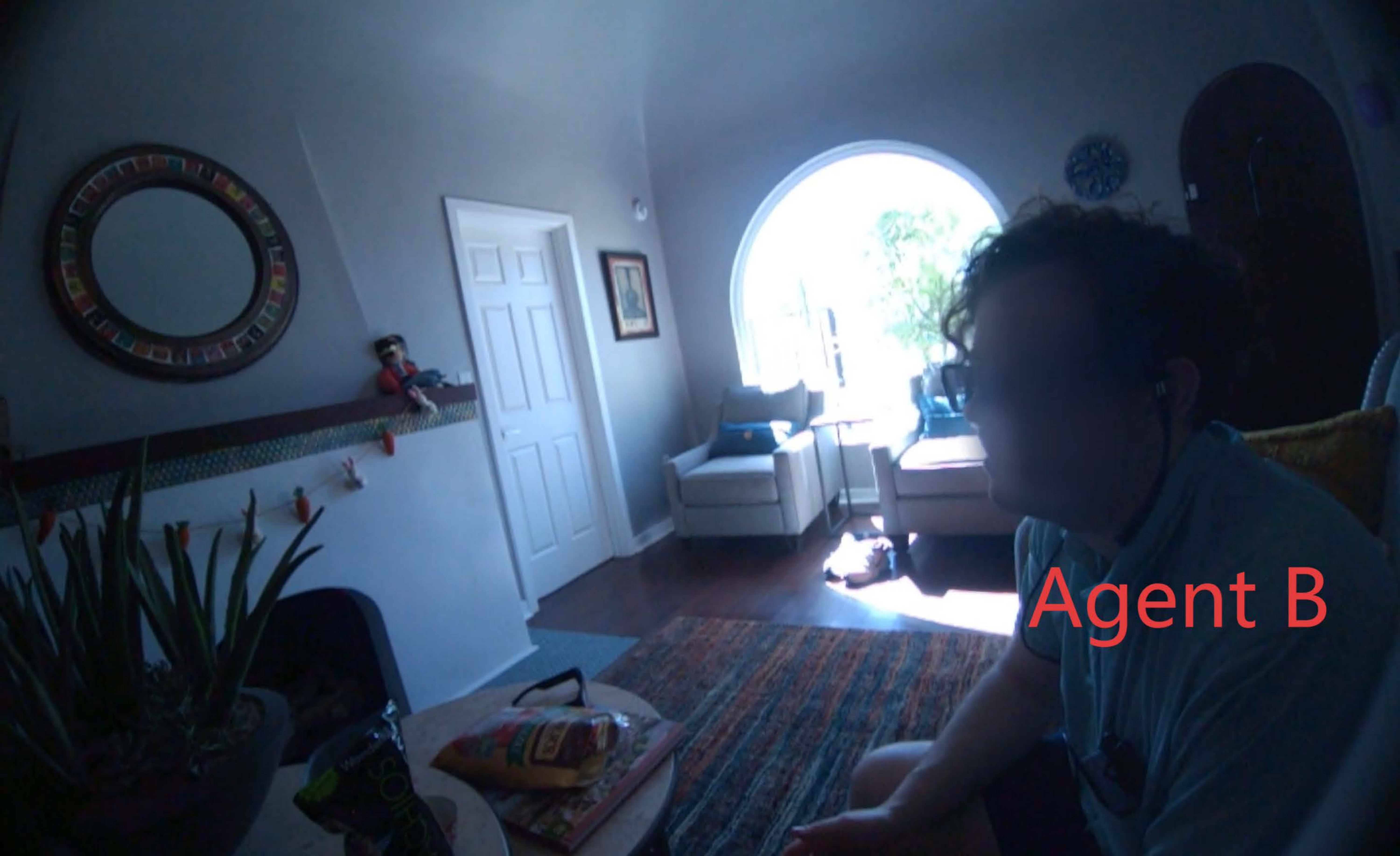} & 
    \includegraphics[width=\linewidth]{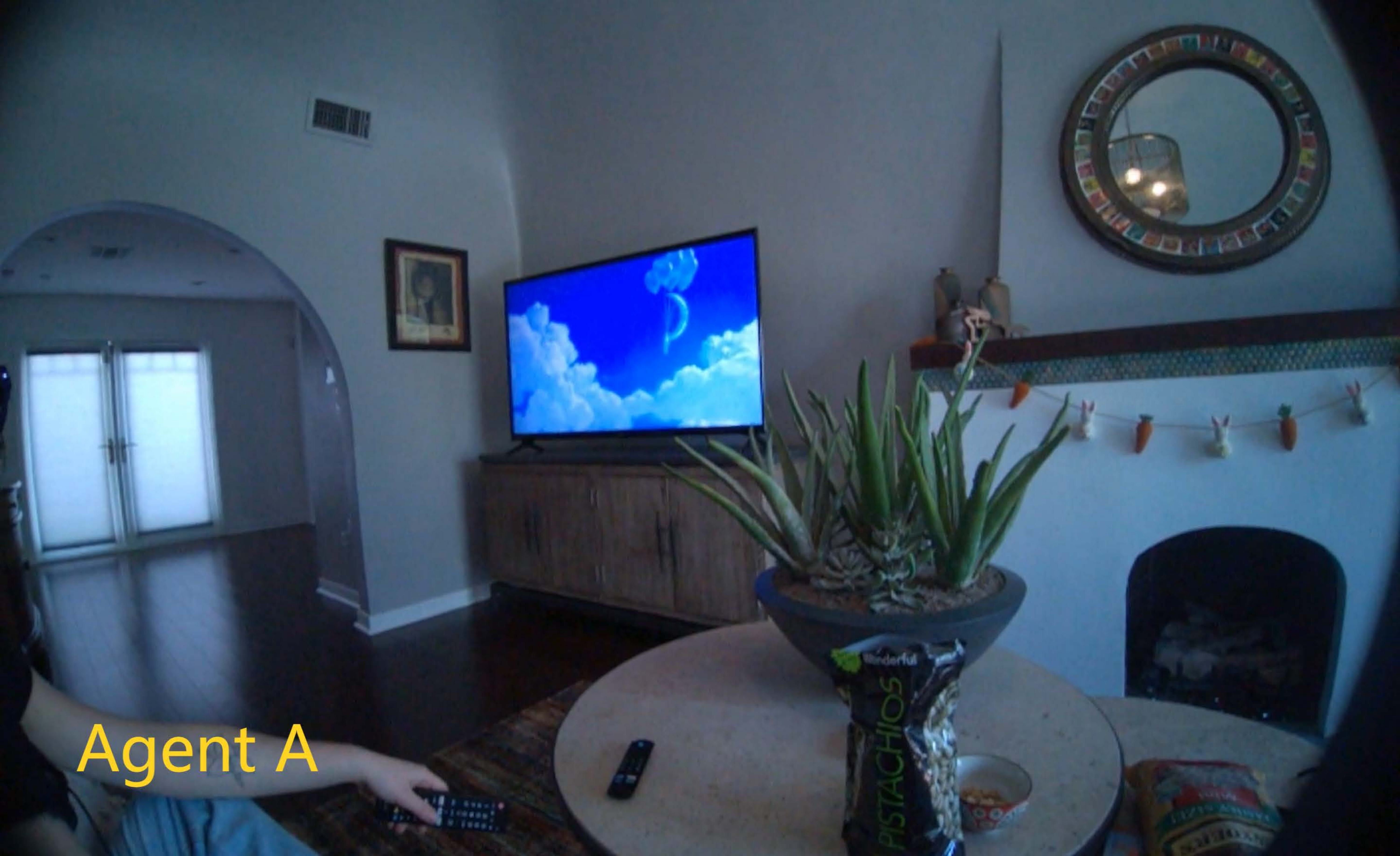} & 
    \scriptsize
    \textbf{Baseline 1 (Ego):} \newline
    front-right \ding{55}\newline
    \textit{Error:} Just copies what Agent A sees and forgets to flip left and right.\newline
    \textbf{Ours:} front-left \ding{51}\newline
    \textit{Reasoning:} Figures out which way Agent B is facing and correctly rotates the view. \\
    \midrule
    
    \textit{Mutually Invisible} \newline (Case 1) & 
    \includegraphics[width=\linewidth]{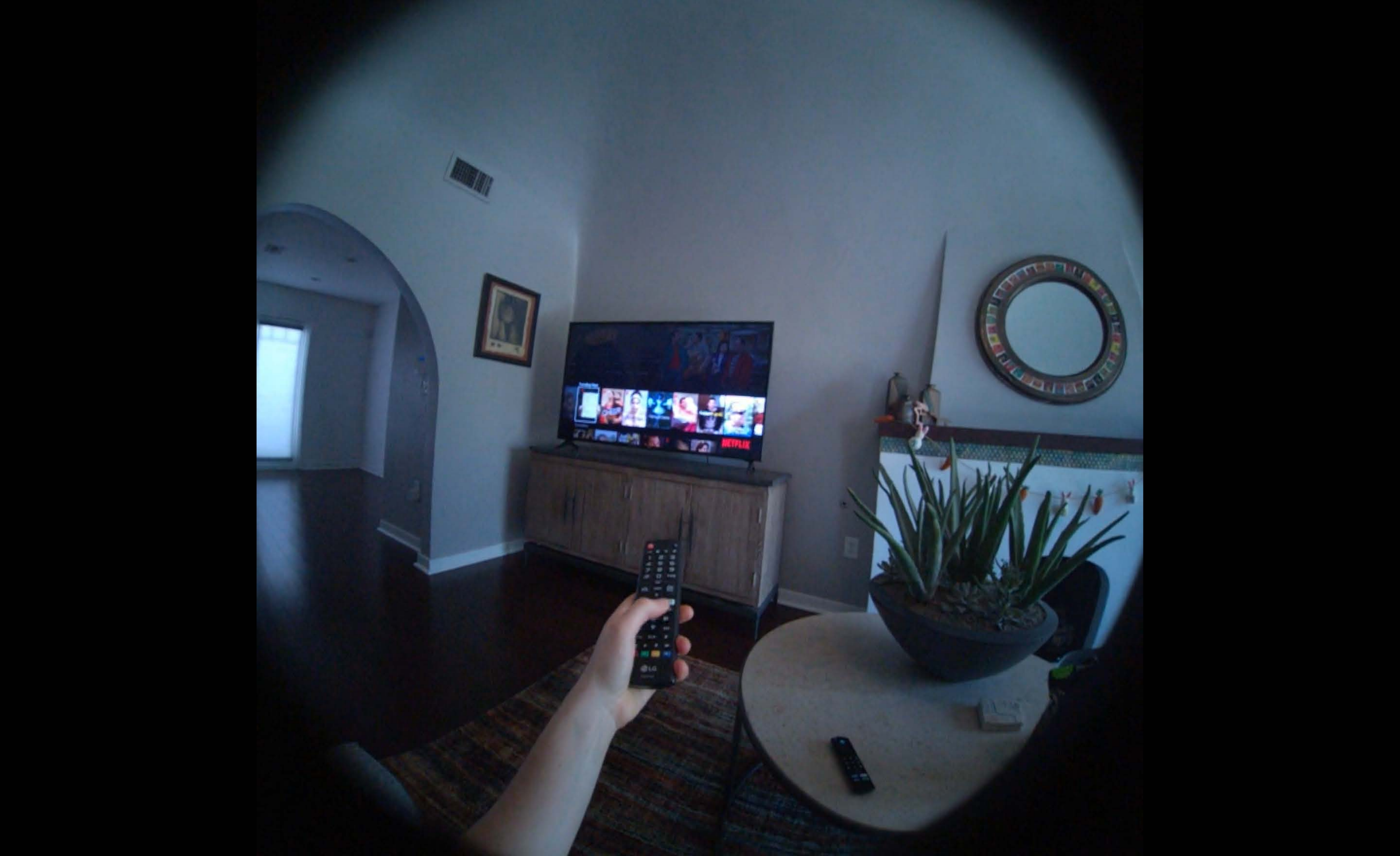} & 
    \includegraphics[width=\linewidth]{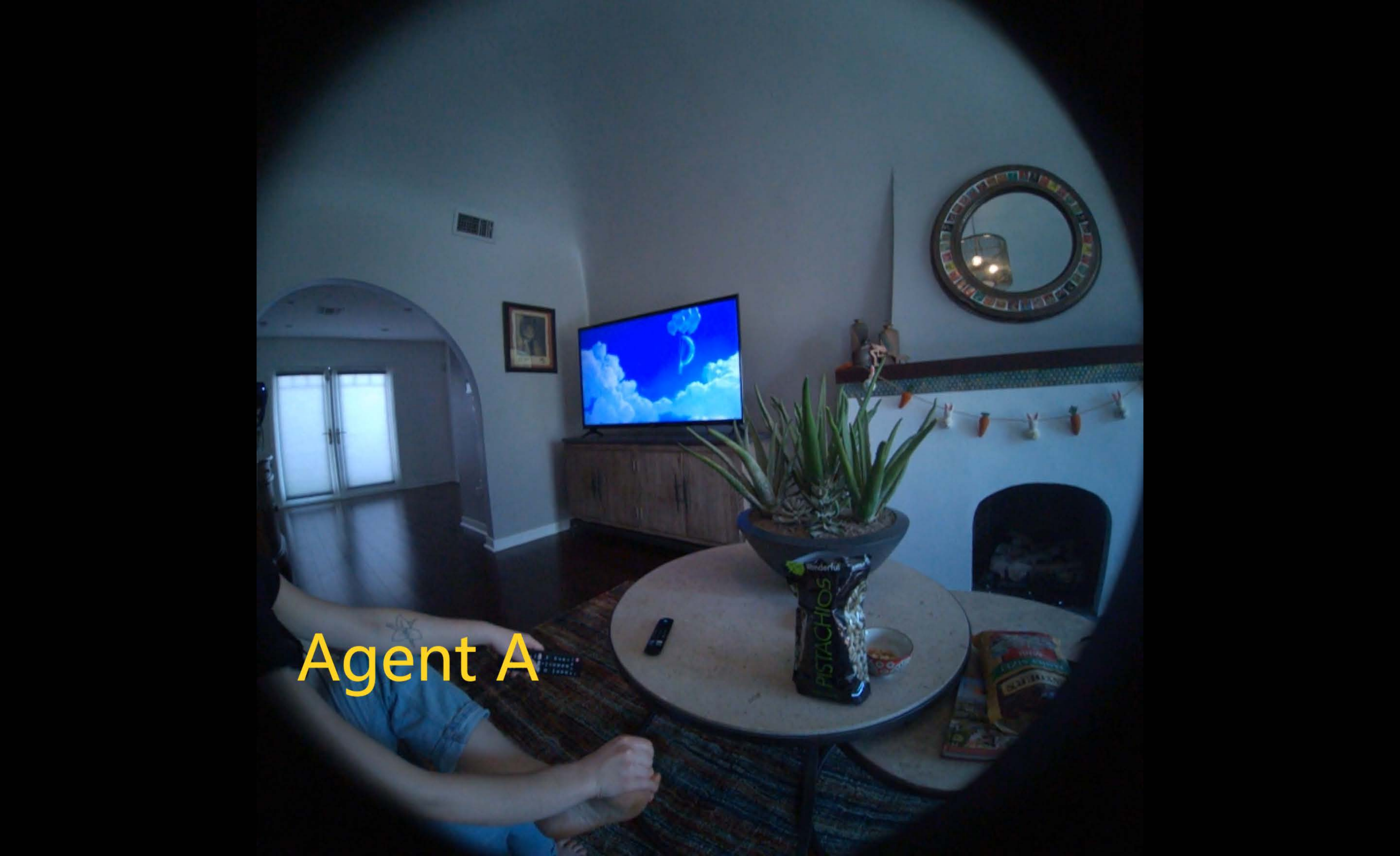} & 
    \scriptsize
    \textbf{Baseline 1 (Ego):} \newline
    back-right \ding{55}\newline
    \textit{Error:} Completely lost without visual clues, just makes a random guess.\newline
    \textbf{Ours:} front-left \ding{51}\newline
    \textit{Reasoning:} Knows it can't see anything, so it smartly uses sound to locate Agent A. \\
    \midrule
    
    \textit{Mutually Invisible} \newline (Case 2) & 
    \includegraphics[width=\linewidth]{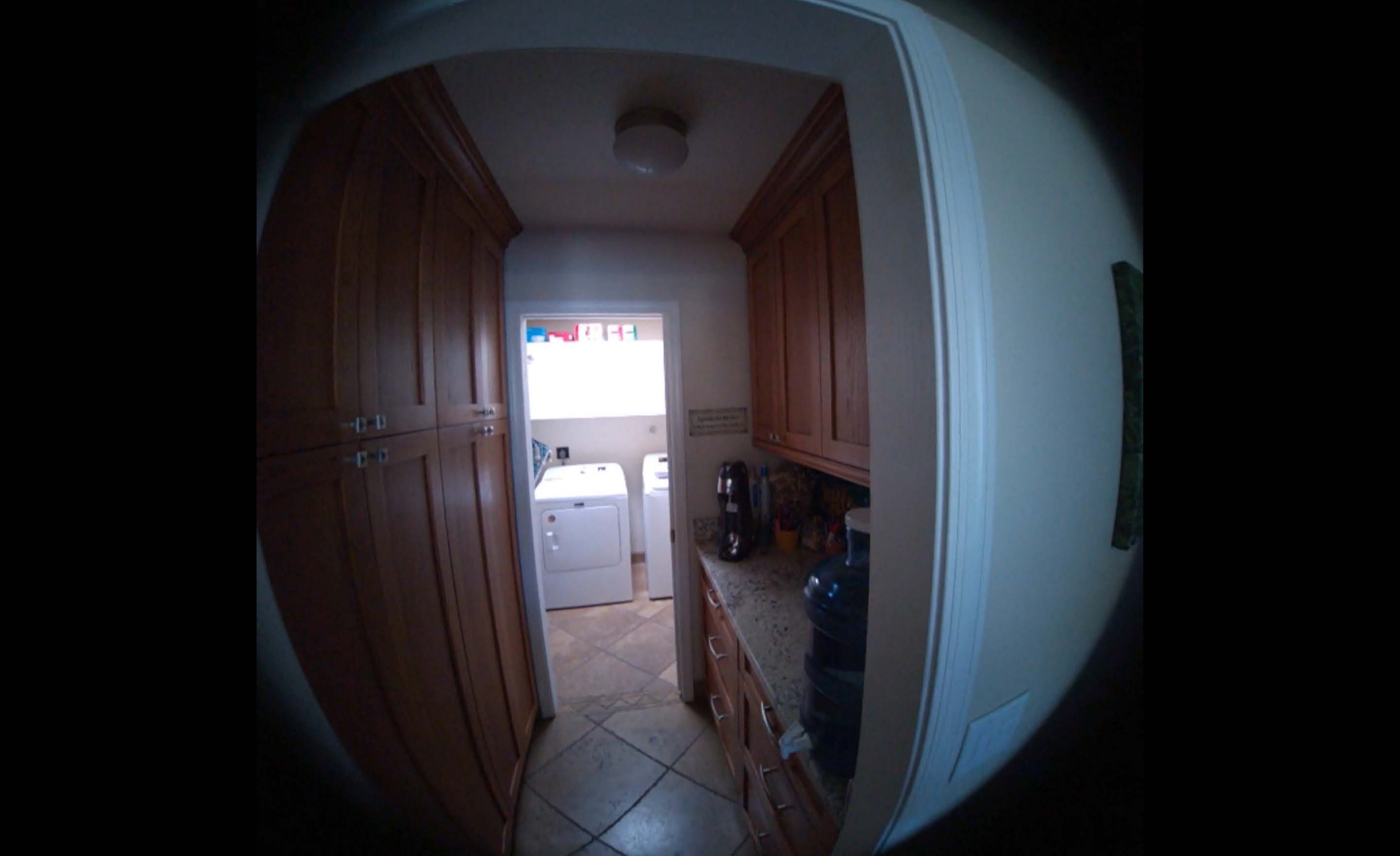} & 
    \includegraphics[width=\linewidth]{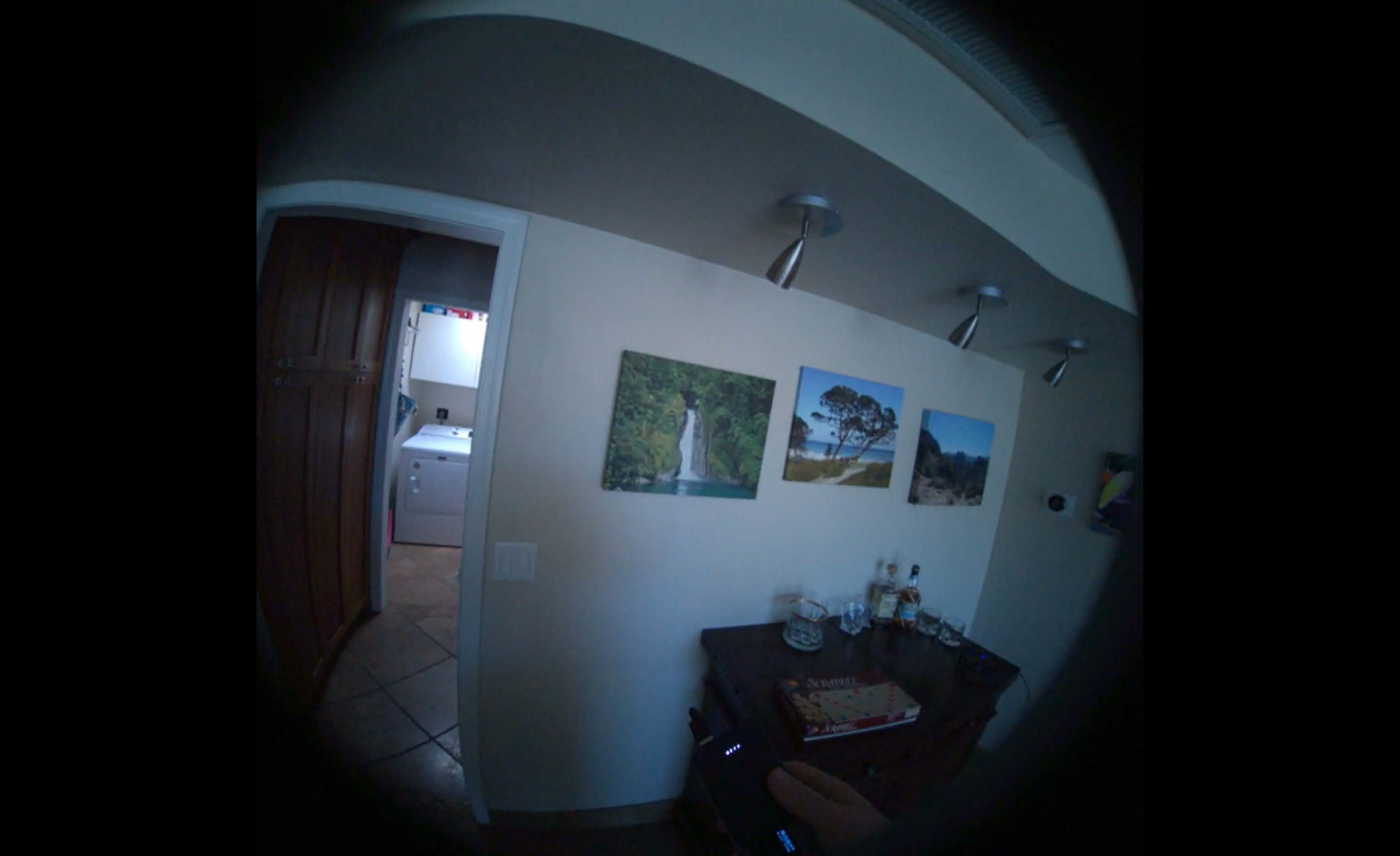} & 
    \scriptsize
    \textbf{Baseline 1 (Ego):} \newline
    back-right \ding{55}\newline
    \textit{Error:} Gives up and guesses blindly because both agents are hidden.\newline
    \textbf{Ours:} front-left \ding{51}\newline
    \textit{Reasoning:} Stops guessing visually and uses clear audio clues to find the correct position. \\
    \midrule
    
    \textit{Mutually Invisible} \newline (Case 3) & 
    \includegraphics[width=\linewidth]{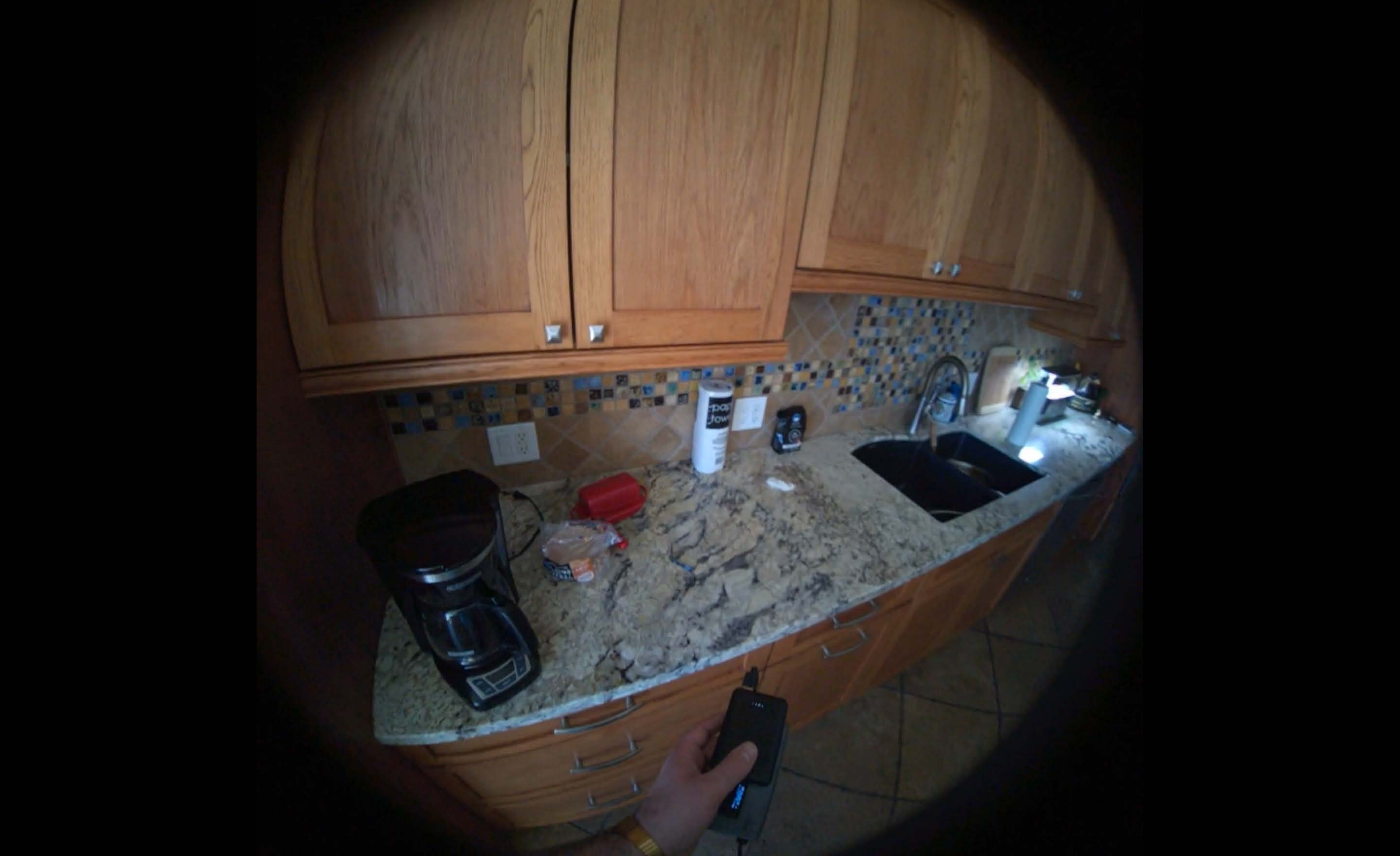} & 
    \includegraphics[width=\linewidth]{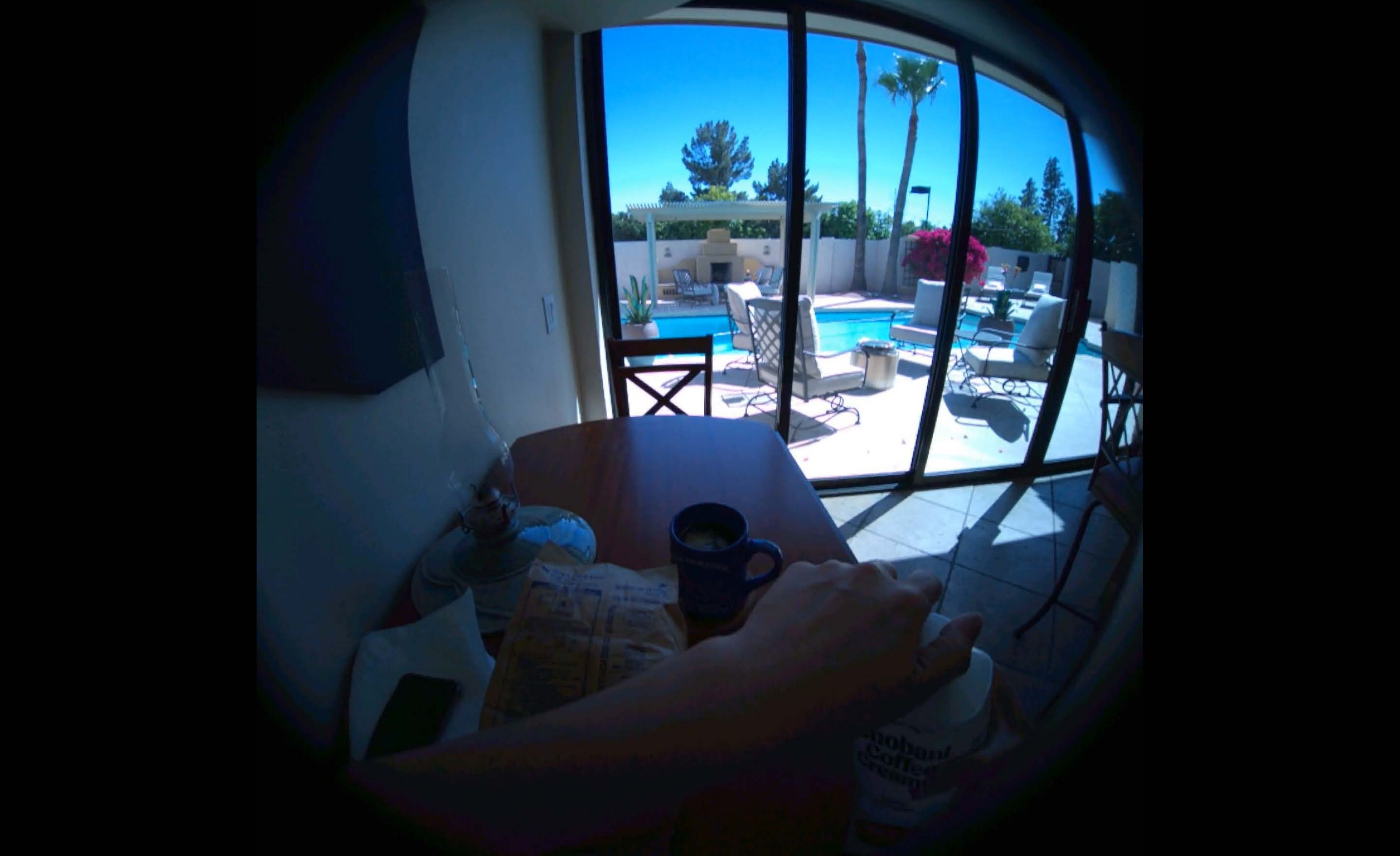} & 
    \scriptsize
    \textbf{Baseline 1 (Ego):} \newline
    back-left \ding{55}\newline
    \textit{Error:} Gets confused about left and right, picking the wrong side.\newline
    \textbf{Ours:} back-right \ding{51}\newline
    \textit{Reasoning:} Uses sound to tell left from right, maintaining a strong sense of direction. \\
    \bottomrule
    \end{tabular}
    
    \caption{Comprehensive Qualitative Analysis: we contrast the reasoning outputs of the End-to-End baseline versus our Observe-to-Believe pipeline across $4$ distinct spatial configurations. 
    Our method successfully navigates epistemic bottlenecks (\texttt{AOnlySeeB} and \texttt{Mutually Invisible}) where traditional visual-language models experience severe Cartesian Illusions or catastrophic failure.}
    \label{tab:all_typical_examples}
\end{table*}

\subsection{Qualitative Analysis \& Typical Examples}
\label{subsec:typical_examples}

To provide deeper intuition into the cognitive dynamics of our \textit{Observe-to-Believe} pipeline, we present a qualitative comparison featuring four representative scenarios from the \texttt{AOnlySeeB} and \texttt{Mutually Invisible} subsets (Table \ref{tab:all_typical_examples}). These examples systematically expose the fundamental failure modes of end-to-end reasoning and demonstrate how explicit spatial horizon conversion resolves them.

\paragraph{Breaking the Cartesian Illusion (\texttt{AOnlySeeB}).}
As illustrated in Table \ref{tab:all_typical_examples}(a), asymmetrical visibility presents a severe cognitive trap for traditional Vision-Language Models. In this scenario, Agent B is highly visible within Agent A's egocentric camera, yet A remains completely outside of B's field of view. Baseline 1 falls victim to the \textit{Cartesian Illusion}: it leverages the visual prominence of B in the input frame to hallucinate a shared visual horizon, predicting erroneous forward-facing coordinates. 

Conversely, our Stage I sensory cortex correctly extracts structural cues (e.g., Agent B's back is facing the camera, yielding a specific $\hat{\theta}_B$), leading the pipeline to actively enforce the spatial horizon mask ($M_v = 0$). By explicitly determining that visual priors are invalid for Agent B, the Stage II reasoning mechanism forcefully routes the context to non-visual spatial relations (such as audio cues), completely avoiding the spatial flip curse across various relative orientations.

\paragraph{Navigating the Epistemic Bottleneck (\texttt{Mutually Invisible}).}
Table \ref{tab:all_typical_examples}(b) showcases the extreme boundary of Theory of Mind reasoning, where the visual stream collapses entirely for both agents. In these instances, Baseline 1 devolves into pure statistical guesswork, outputting arbitrary spatial coordinates because it lacks the capacity to decouple visual absence from spatial existence. 

Our pipeline maintains high-fidelity reasoning under complete occlusion. Because the Stage I observation extraction returns no visual bounding boxes for the target agent, $M_v$ is definitively set to $0$. The explicit textual reasoning trace of our model clearly articulates this epistemic bottleneck, deliberately shifting its cognitive reliance to spatial audio directionality and environmental descriptions. This consistent cross-modal routing empirically proves that higher-order spatial reasoning in complex environments necessitates a decoupled, explicitly routed cognitive architecture rather than implicit, end-to-end embedding fusion.

\section{Conclusion}
\label{sec:conclusion}
This work introduces the \textbf{Observe-to-Believe} pipeline, a framework designed to overcome the ``Cartesian Illusion'' in embodied Theory of Mind. By decoupling raw multi-modal perception from logical spatial reasoning, we demonstrate that recursive belief modeling is most effective when grounded in explicit physical sensory bottlenecks. Our results confirm that modeling the \textit{spatial horizon}—specifically the transition from visual to audio-spatial cues during occlusion—provides a robust pathway for agents to navigate complex, asymmetrical multi-agent environments. This architecture effectively resolves the ``spatial flip curse'' and offers a scalable template for integrating modality-aware perspective shifting into future autonomous systems.

\section{Limitations and Future Work}
\label{sec:limitations}
Despite its improved accuracy, our pipeline exhibits a performance-latency trade-off, where the comprehensive Stage I perceptual extraction accounts for the majority of the computational overhead. Furthermore, our current implementation relies on a discretized 8-direction spatial model; while sufficient for high-level ToM reasoning, it lacks the resolution required for fine-grained continuous control. Future research will explore \textbf{Active Perception}, allowing Stage II reasoning to dynamically request higher-resolution perceptual re-scans when epistemic uncertainty is detected. Additionally, we plan to generalize this framework to $N$-order ToM scenarios, enabling agents to model nested beliefs (e.g., ``A thinks that B knows that A is hidden'') in adversarial or high-stakes cooperative settings.

\section{Broader Impacts and Ethical Considerations}

The advancement of computational Second-Order Theory of Mind (ToM) and modality-aware spatial reasoning through frameworks like our proposed Observe-to-Believe pipeline offers profoundly positive implications for the future of Embodied AI and human-robot interaction. By explicitly breaking the ``Cartesian Illusion'' and enabling machines to understand exactly what a human can or cannot perceive, this work paves the way for deeply cooperative and safety-conscious systems. In assistive robotics, for instance, machines could proactively adapt their communication or actions to accommodate users with restricted fields of view or sensory impairments. In autonomous driving, vehicles could more accurately deduce a pedestrian's situational awareness, drastically reducing accidents in shared spaces. Furthermore, search-and-rescue systems could leverage our modality-routing mechanisms to navigate hazardous, occluded environments and locate survivors using spatial audio. While the ability to explicitly compute an individual's visual and auditory horizons necessitates thoughtful privacy safeguards to prevent unintended surveillance or boundary overreach, formalizing these bottlenecks transparently allows the research community to proactively design robust alignment frameworks. Ultimately, our work provides a critical, foundational step toward empathetic AI systems that naturally respect human perceptual limitations and work seamlessly to enhance real-world safety and well-being.

\begin{ack}
This work was supported by NSFC-62406034.
\end{ack}






\bibliographystyle{plainnat}
\bibliography{reference}
\clearpage


\appendix

\section{Technical appendices and supplementary material}



\begin{tcolorbox}[
    colframe=gray!60!black,       
    colback=white,                
    colbacktitle=gray!15!white,   
    coltitle=black,               
    fonttitle=\bfseries,          
    title={Prompt: Video Analysis (gemini-2.5-pro)}, 
    boxrule=0.5pt,                
    arc=1mm,                      
    left=2mm, right=2mm, top=2mm, bottom=2mm
]
Analyze the current input video.\\
Identify Agent B (the other person) relative to Agent A (the camera wearer).

\vspace{1ex}
\noindent\textbf{[Task]}
\begin{itemize}
    \setlength{\itemsep}{0pt}
    \setlength{\parskip}{0pt}
    \item Track Agent B across the current input only.
    \item When Agent B is clearly visible, you should output a key frame for that moment even if no large state change happens.
\end{itemize}

\vspace{1ex}
\noindent\textbf{[Field Rules]}
\begin{itemize}
    \setlength{\itemsep}{0pt}
    \setlength{\parskip}{0pt}
    \item \texttt{is\_static}: true if Agent B is roughly stable.
    \item \texttt{distance}: estimated meters from Agent A to Agent B.
    \item \texttt{direction}: estimated egocentric signed horizontal angle from Agent A to Agent B. positive (\texttt{+}) values when Agent B is to the right of Agent A.
    \item \texttt{b\_orientation\_to\_camera} must be exactly one of \texttt{front-left}, \texttt{front-right}, \texttt{back-left}, \texttt{back-right}. These values describe which specific quadrant of Agent B's head is directly facing Agent A: \texttt{front-left}: The front-left side of Agent B's head (e.g., their left cheek/eye area) is facing Agent A.
    \item Human face/head turning is usually limited to a small range around straight ahead, and most of the time stays within about +/-30 degrees rather than extreme rotations.
    \item As a common-sense prior, people usually keep their hands and arms in front of their torso or within a natural side range, and generally do not place both hands behind their back during ordinary interaction.
    \item \texttt{b\_orientation\_confidence} must from \texttt{0.0} to \texttt{1.0}.
    \item \texttt{visibility\_to\_camera}: one of \texttt{visible}, \texttt{occluded}, \texttt{uncertain}.
\end{itemize}

\vspace{1ex}
\noindent\textbf{[Question]}\\
\{\{QUESTION\_TEXT\}\}

\vspace{1ex}
\noindent\textbf{[Context]}\\
{\ttfamily
\{\\
\hspace*{4ex}"start\_time": "m:ss.mmm",\\
\hspace*{4ex}"end\_time": "m:ss.mmm",\\
\hspace*{4ex}"option": "front-left","front-right","back-left","back-right"\\
\}
}

\vspace{1ex}
\noindent\textbf{[Output schema]}\\
{\ttfamily
\{\\
\hspace*{4ex}"key\_frames": \{\\
\hspace*{8ex}"TIMESTAMP": \{\\
\hspace*{12ex}"is\_static": boolean,\\
\hspace*{12ex}"distance": "string\_meters",\\
\hspace*{12ex}"direction": "string\_degrees",\\
\hspace*{12ex}"b\_orientation\_to\_camera": "string",\\
\hspace*{12ex}"b\_orientation\_confidence": float,\\
\hspace*{12ex}"visibility\_to\_camera": "visible|occluded|uncertain",\\
\hspace*{12ex}"description": \{\\
\hspace*{16ex}"event\_summary": \{ "object\_name": "direction" \}\\
\hspace*{12ex}\}\\
\hspace*{8ex}\}\\
\hspace*{4ex}\}\\
\}
}
\end{tcolorbox}
\nopagebreak
\captionof{figure}{The detailed prompt used for Gemini-2.5-Pro video analysis task.}
\label{fig:prompt_gemini}

\begin{figure}[htbp]
\centering
\begin{tcolorbox}[
    colframe=gray!60!black,       
    colback=white,                
    colbacktitle=gray!15!white,   
    coltitle=black,               
    fonttitle=\bfseries,          
    title={Prompt: Second-Order ToM Reasoning (DeepSeek-V4-Flash)}, 
    boxrule=0.5pt,                
    arc=1mm,                      
    left=2mm, right=2mm, top=2mm, bottom=2mm
]
You are performing benchmark-conditioned high-level ToM reasoning.

\vspace{1ex}
\noindent\textbf{[Goal]}\\
Predict what Agent B believes about Agent A's position during the benchmark clip from \{\{START\_TIME\}\} to \{\{END\_TIME\}\} ('B thinks A is at [X]').

\vspace{1ex}
\noindent\textbf{[Reasoning rules]}
\begin{itemize}
    \setlength{\itemsep}{0pt}
    \setlength{\parskip}{0pt}
    \item \textbf{Second-Order Logic}: Map A's observations to B's perspective. If A sees B facing a certain direction, where does B think A is?
    \item \textbf{Direct Orientation Rule}: If \texttt{visibility\_to\_camera} is \texttt{visible} and B is clearly observable, use the observed \texttt{b\_orientation\_to\_camera} as the final \texttt{belief\_direction}. In this case, the side of B that A sees is the side where A is located relative to B.
    \item \textbf{Joint Recovery Rule}: If A cannot clearly see B (\texttt{visibility\_to\_camera} is not \texttt{visible}), jointly use the structured audio evidence, neighboring visual frames, neighboring frames' \texttt{is\_static}, neighboring frames' B orientation, and Agent A's self-motion from \texttt{a\_world} / \texttt{a\_orientation\_deg} to infer A's position relative to B. If \texttt{is\_static} stays true throughout, prefer keeping B's last reliable orientation-based belief stable unless A's own motion indicates a consistent change.
    \item \textbf{Self-Motion Compensation}: If A moves, distinguish A's own motion from B's motion. Use the motion-compensated B world belief to avoid confusing ego-rotation with B movement.
    \item \textbf{Audio-Motion Coupling}: When audio bearing or energy changes while A moves, use that coupling as a consistency check on B's world-frame location.
\end{itemize}

\vspace{1ex}
\noindent Return strict JSON with exactly this schema and no other keys\\
{\ttfamily
\{\\
"belief\_direction": "front-left","front-right","back-left","back-right"\\
\}
}

\vspace{1ex}
\noindent\textbf{[A sees B]}\\
{\ttfamily
\{\\
\hspace*{4ex}"start\_time": "m:ss.mmm",\\
\hspace*{4ex}"end\_time": "m:ss.mmm",\\
\hspace*{4ex}"a\_world\_at\_clip\_end": [x, y, z],\\
\hspace*{4ex}"a\_orientation\_deg\_at\_clip\_end": degrees,\\
\hspace*{4ex}"visual\_evidence": \{\\
\hspace*{8ex}"key\_frames": \{\\
\hspace*{12ex}"TIMESTAMP": \{\\
\hspace*{16ex}"\{\{ Stage 1 (Gemini) output fields \}\}",\\
\hspace*{16ex}"a\_world": [x, y, z],\\
\hspace*{16ex}"a\_orientation\_deg": degrees\\
\hspace*{12ex}\}\\
\hspace*{8ex}\}\\
\hspace*{4ex}\}\\
\}
}

\vspace{1ex}
\noindent\textbf{[Structured audio evidence]}\\
{\ttfamily
\{\\
\hspace*{4ex}"audio\_features": \{ ... \},\\
\hspace*{4ex}"audio\_summary": \{ ... \},\\
\hspace*{4ex}"spatial\_fps": int\\
\}
}
\end{tcolorbox}
\caption{The detailed prompt used for DeepSeek-V4-Flash in the second-order ToM reasoning stage.}
\label{fig:prompt_deepseek}
\end{figure}

\subsection{Detailed Prompting Templates for the Hierarchical ToM Pipeline}

In our hierarchical reasoning framework for multi-modal Second-Order Theory of Mind (ToM) tasks, we utilize two distinct large language models tailored to their specific sub-tasks. 

\textbf{Stage 1: Ego-view Video Analysis.} For the initial visual perception and agent tracking stage, we employ the \textbf{Gemini-2.5-Pro} model. This model analyzes the input video to extract frame-level structured evidence regarding the agents' spatial states, orientations, and visibilities. The exact prompt provided to Gemini-2.5-Pro is detailed in Figure~\ref{fig:prompt_gemini}.

\textbf{Stage 2: Second-Order Belief Reasoning.} Building upon the perceptual evidence extracted in the first stage, along with structured audio data, we utilize the \textbf{DeepSeek-V4-Flash} model for the high-level cognitive reasoning stage. This model infers the target agent's belief state based on specific logic rules (e.g., direct orientation, joint recovery, and self-motion compensation). The comprehensive prompt used for DeepSeek-V4-Flash is presented in Figure~\ref{fig:prompt_deepseek}.

\subsubsection{Qualitative Analysis of the Observe-to-Believe Pipeline}
\label{sec:appendix_qualitative}

In this section, we provide a detailed breakdown of four representative samples from the SAVVY \textit{ego\_direction} subset to illustrate the internal cognitive dynamics of our \textbf{Observe-to-Believe} pipeline. These cases demonstrate how the decoupled architecture handles various levels of perceptual accuracy and logical complexity across different visibility conditions.

\subsubsection{Case 1: Fully Correct Prediction (S1 \ding{51}, S2 \ding{51}, B1 \ding{51})}
\textbf{Visibility Rule:} \texttt{Multually Visible} \quad \textbf{Ground Truth (Gold):} \texttt{front-left}

\begin{figure}[htbp]
    \centering
    \begin{minipage}{0.48\textwidth}
        \centering
        \includegraphics[width=\linewidth]{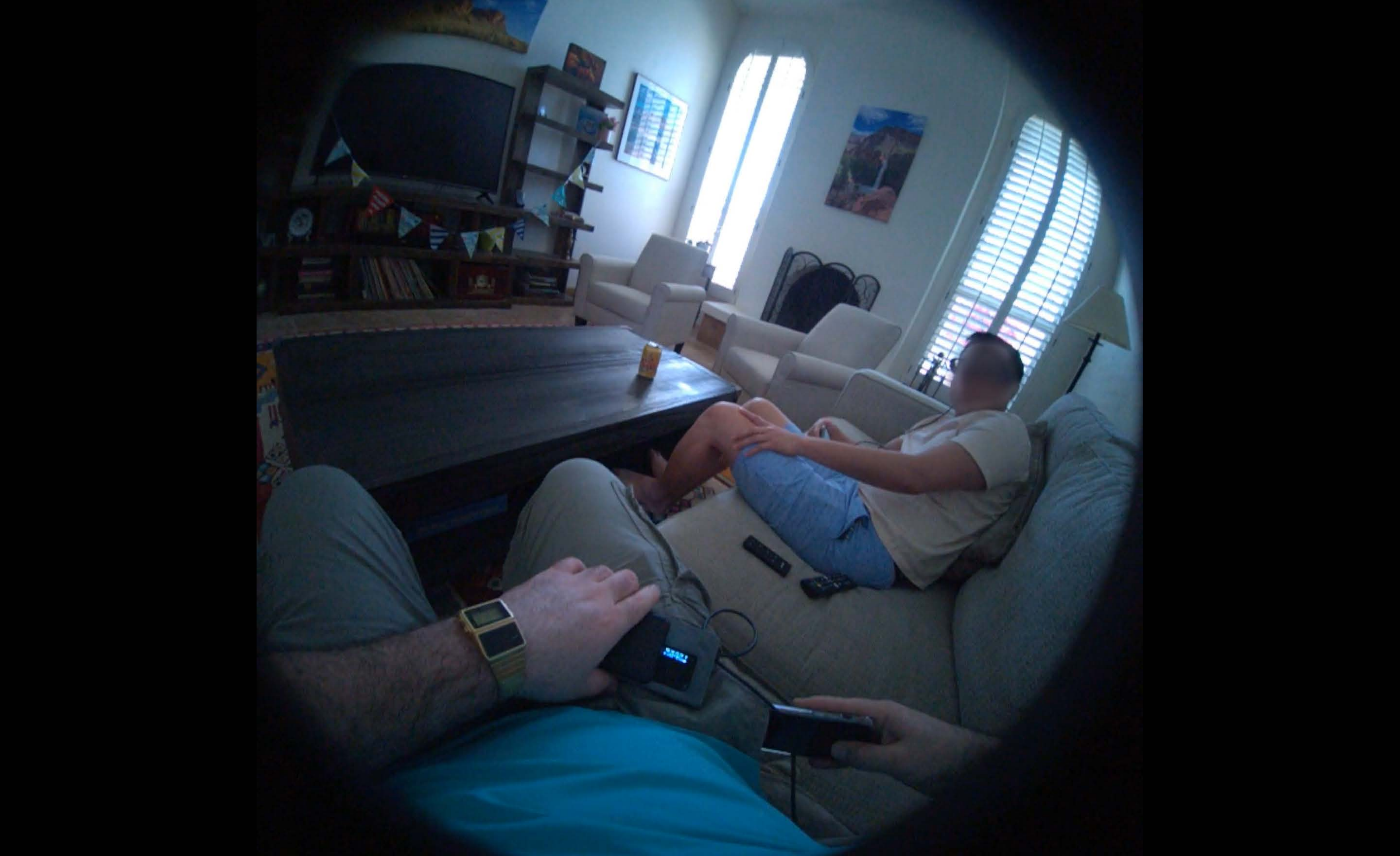}
    \end{minipage}
    \hfill
    \begin{minipage}{0.48\textwidth}
        \centering
        \includegraphics[width=\linewidth]{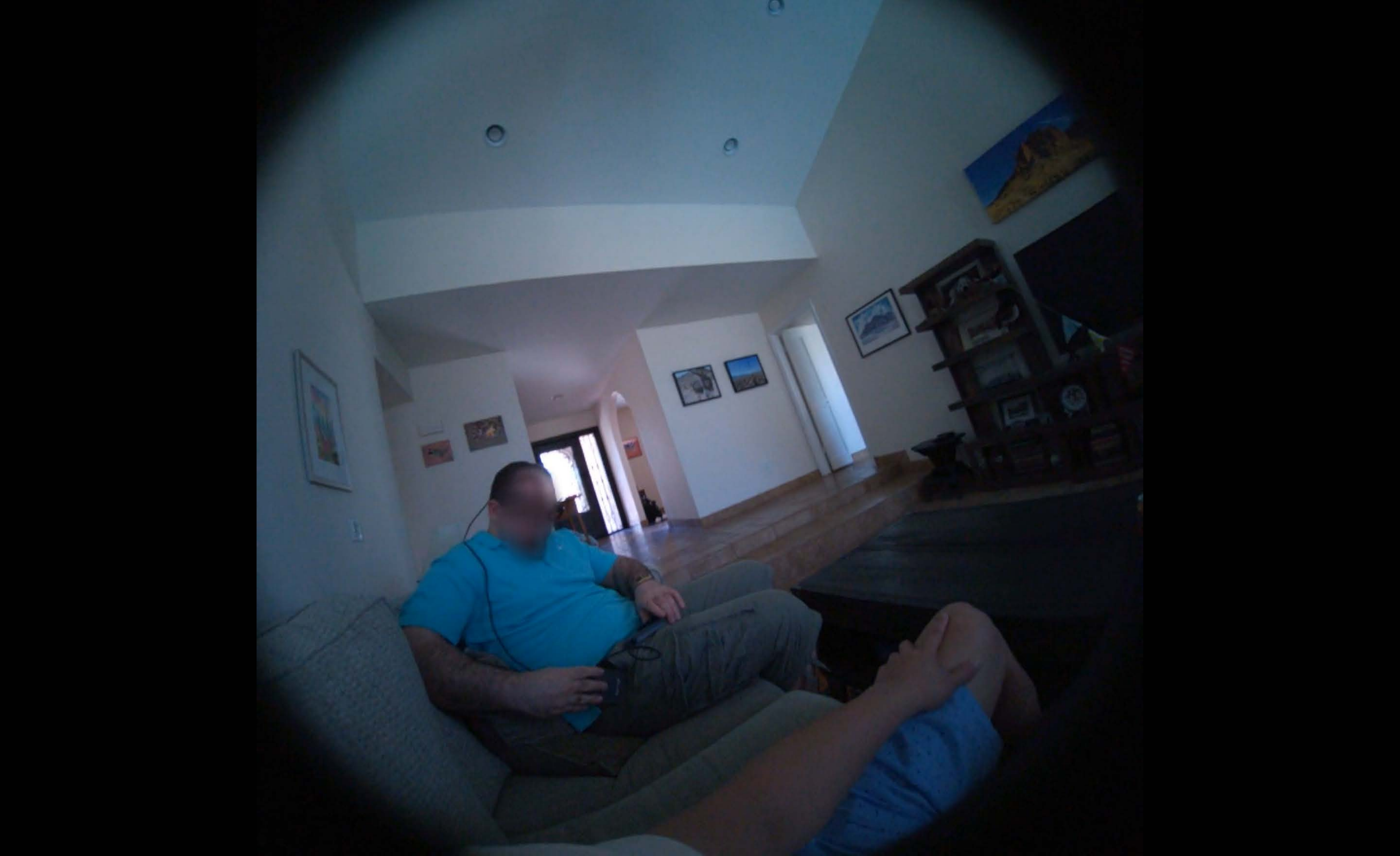}
    \end{minipage}
    \caption{Visual representation of Case 1: All modules maintain high-fidelity reasoning under a shared visual field.}
    \label{fig:case1}
\end{figure}

\textbf{Reasoning Process:}As shown in Figure~\ref{fig:case1}, 1. At the beginning of the interaction (0:00.671--0:04.991), the person is visibly sitting on the couch requesting a tour of the place, positioned near the coffee table and armchair. 2. The Stage 1 perception module processes these keyframes and correctly extracts the agent's orientation as \texttt{front-left}. 3. Relying on the end-to-end mapping, the Baseline 1 model processes this spatial layout and correctly predicts \texttt{front-left}, successfully matching the ground truth. 4. Similarly, the Stage 2 reasoning module leverages the accurate perceptual data to confirm the spatial belief, resulting in a correct final reasoning output of \texttt{front-left}.

\subsubsection{Case 2: Mapping Logic Bias Correction (S1 \ding{51}, S2 \ding{51}, B1 \ding{55})}
\textbf{Visibility Rule:} \texttt{BOnlySeeA} \quad \textbf{Ground Truth (Gold):} \texttt{front-left}

\begin{figure}[htbp]
    \centering
    \begin{minipage}{0.48\textwidth}
        \centering
        \includegraphics[width=\linewidth]{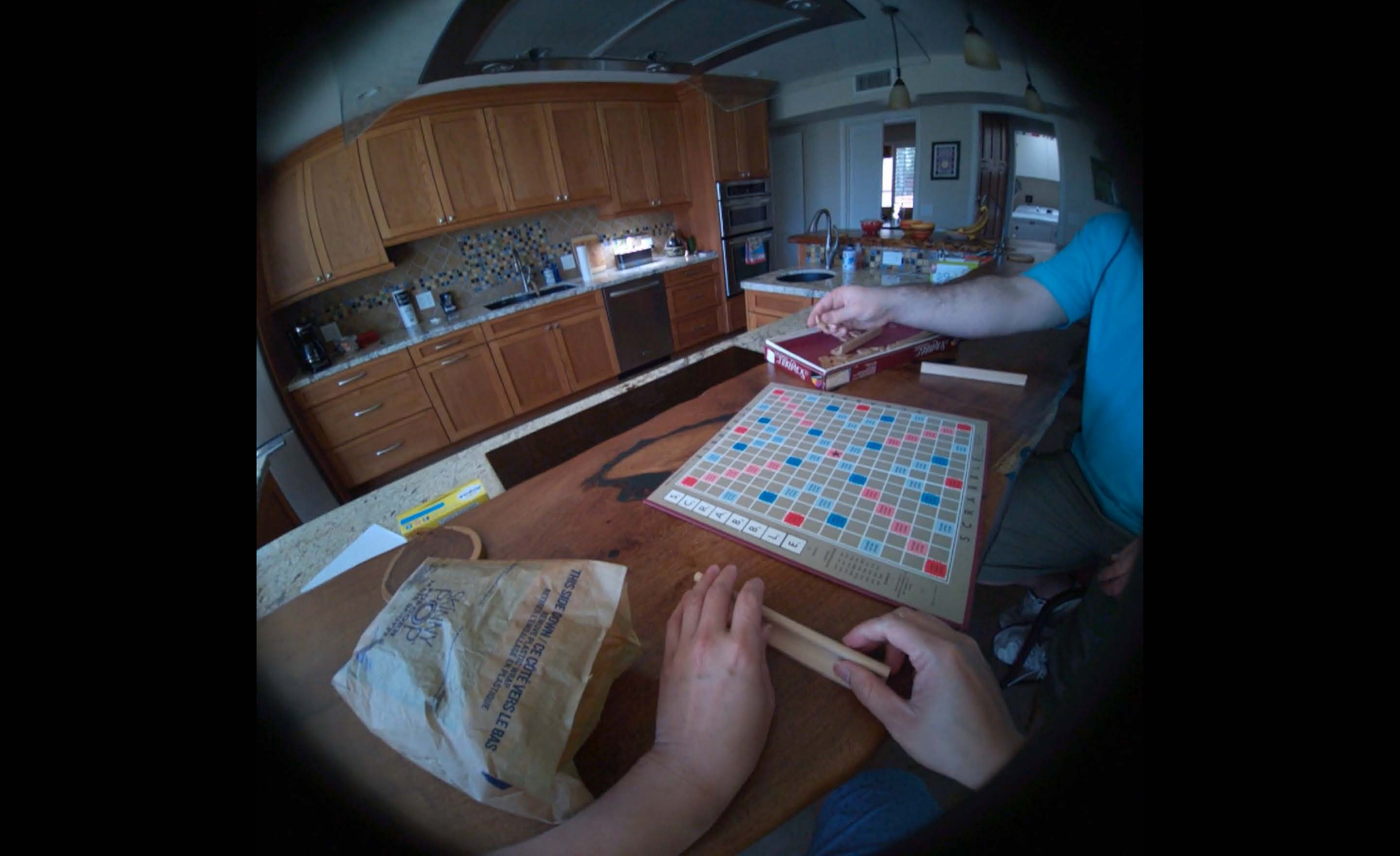}
    \end{minipage}
    \hfill
    \begin{minipage}{0.48\textwidth}
        \centering
        \includegraphics[width=\linewidth]{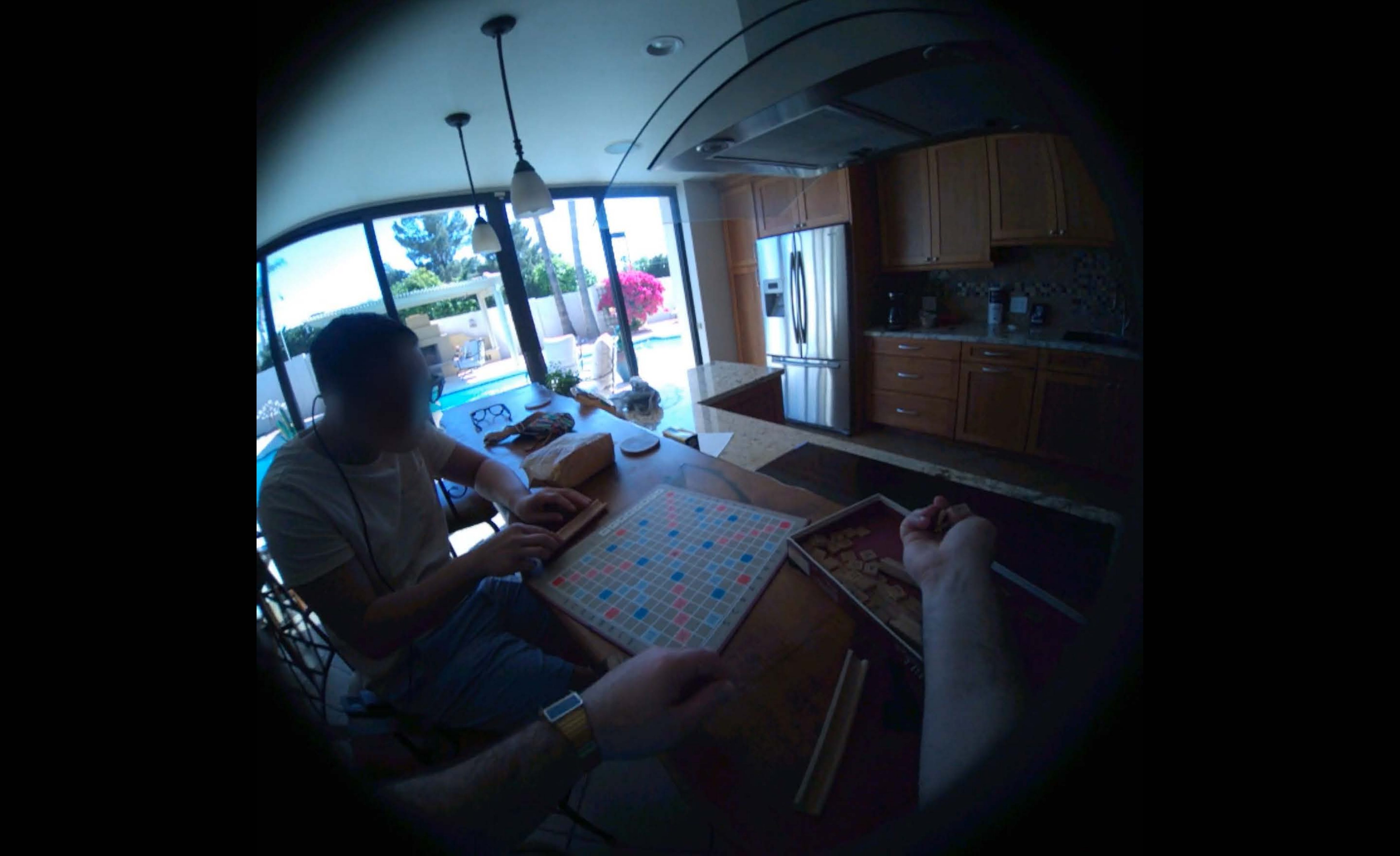}
    \end{minipage}
    \caption{Visual representation of Case 2: Correction of mapping bias where perception is correct but end-to-end mapping fails.}
    \label{fig:case2}
\end{figure}

\textbf{Reasoning Process:}As shown in Figure~\ref{fig:case2}, 1. During the interval 0:28.50--0:31.01, the target agent is actively moving within the camera's field of view. 2. The Stage 1 perception layer accurately processes the visual features and correctly identifies the agent's orientation as \texttt{front-left}. 3. Despite the correct perceptual input, the Baseline 1 end-to-end model experiences a spatial coordinate mapping failure, thereby incorrectly predicting \texttt{front-right}. 4. The Stage 2 reasoning module, however, robustly interprets the raw perceptual evidence and successfully corrects this mapping bias, yielding the correct final reasoning output of \texttt{front-left}.

\subsubsection{Case 3: Reasoning Overriding Perceptual Error (S1 \ding{55}, S2 \ding{51}, B1 \ding{55})}
\textbf{Visibility Rule:} \texttt{BOnlySeeA} \quad \textbf{Ground Truth (Gold):} \texttt{back-right}

\begin{figure}[htbp]
    \centering
    \begin{minipage}{0.48\textwidth}
        \centering
        \includegraphics[width=\linewidth]{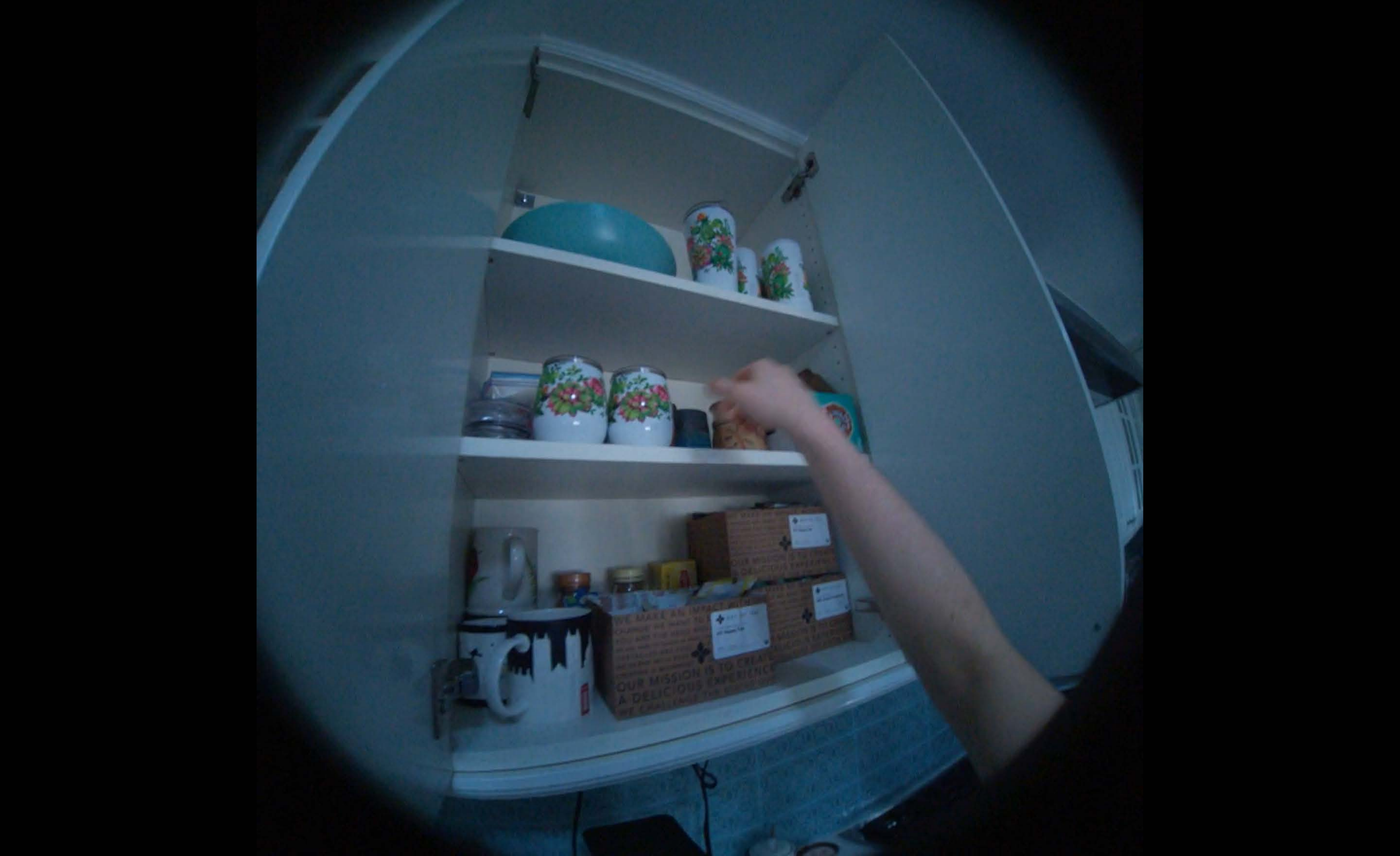}
    \end{minipage}
    \hfill
    \begin{minipage}{0.48\textwidth}
        \centering
        \includegraphics[width=\linewidth]{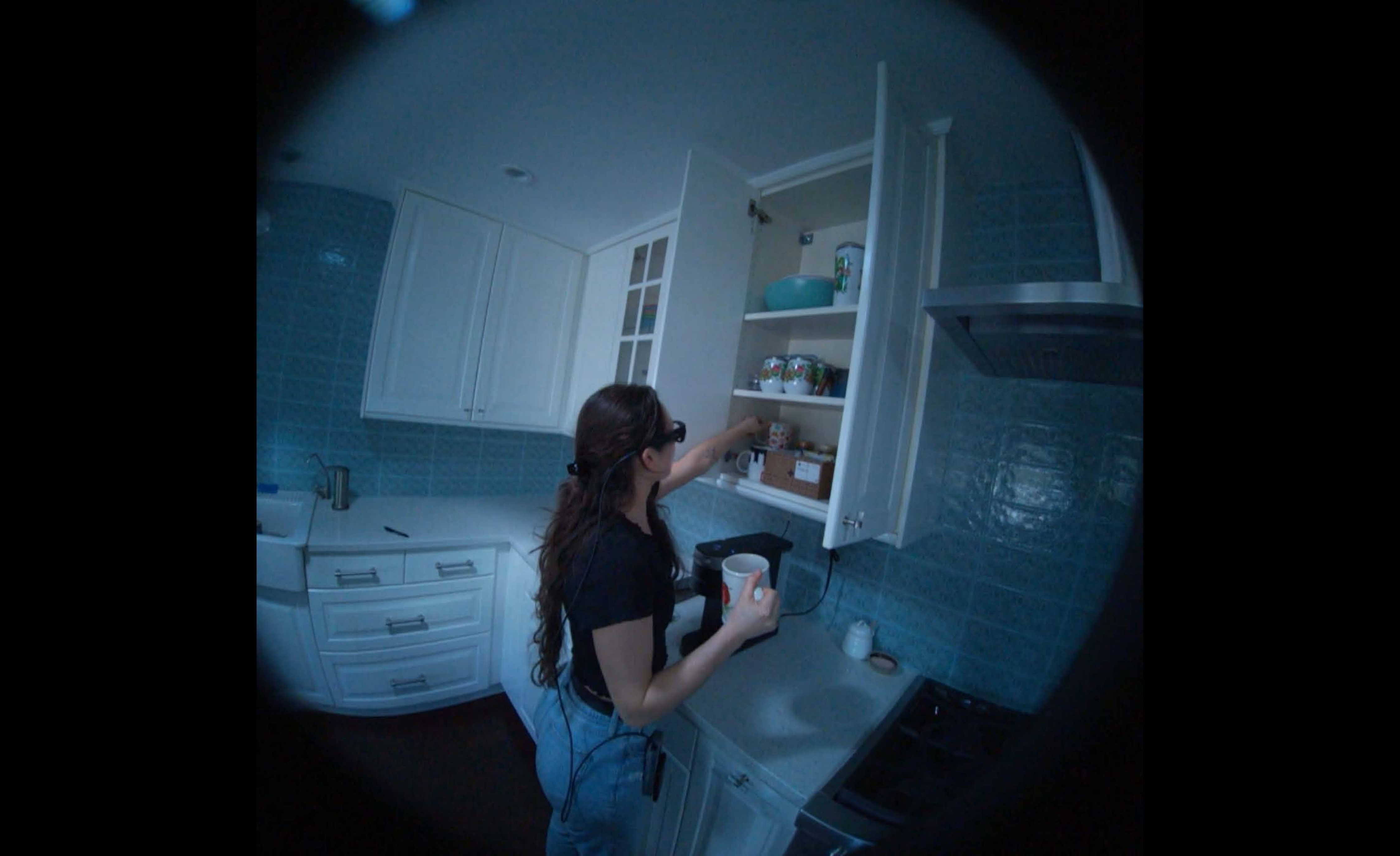}
    \end{minipage}
    \caption{Visual representation of Case 3: Stage 2 overrides a flawed perception by leveraging logical and environmental context.}
    \label{fig:case3}
\end{figure}

\textbf{Reasoning Process:}As shown in Figure~\ref{fig:case3}, 1. Starting at 0:06.666, the agent turns and leaves the room, subsequently choosing a mug near the kitchen cabinet at 0:31.251, becoming heavily occluded from the camera's view. 2. Deprived of clear visual features due to this occlusion, the Stage 1 perception module incorrectly misjudges the target's orientation as \texttt{front-right}. 3. Misled by this visual ambiguity and flawed feature extraction, the Baseline 1 model incorrectly predicts \texttt{front-right}. 4. Conversely, the Stage 2 reasoning module leverages prior spatial context and environmental logic to detect the perceptual inconsistency, successfully overriding the perception error to output the correct target position as \texttt{back-right}.

\subsubsection{Case 4: Systematic Error (S1 \ding{55}, S2 \ding{55}, B1 \ding{55})}
\textbf{Visibility Rule:} \texttt{AOnlySeeB} \quad \textbf{Ground Truth (Gold):} \texttt{back-left}

\begin{figure}[htbp]
    \centering
    \begin{minipage}{0.48\textwidth}
        \centering
        \includegraphics[width=\linewidth]{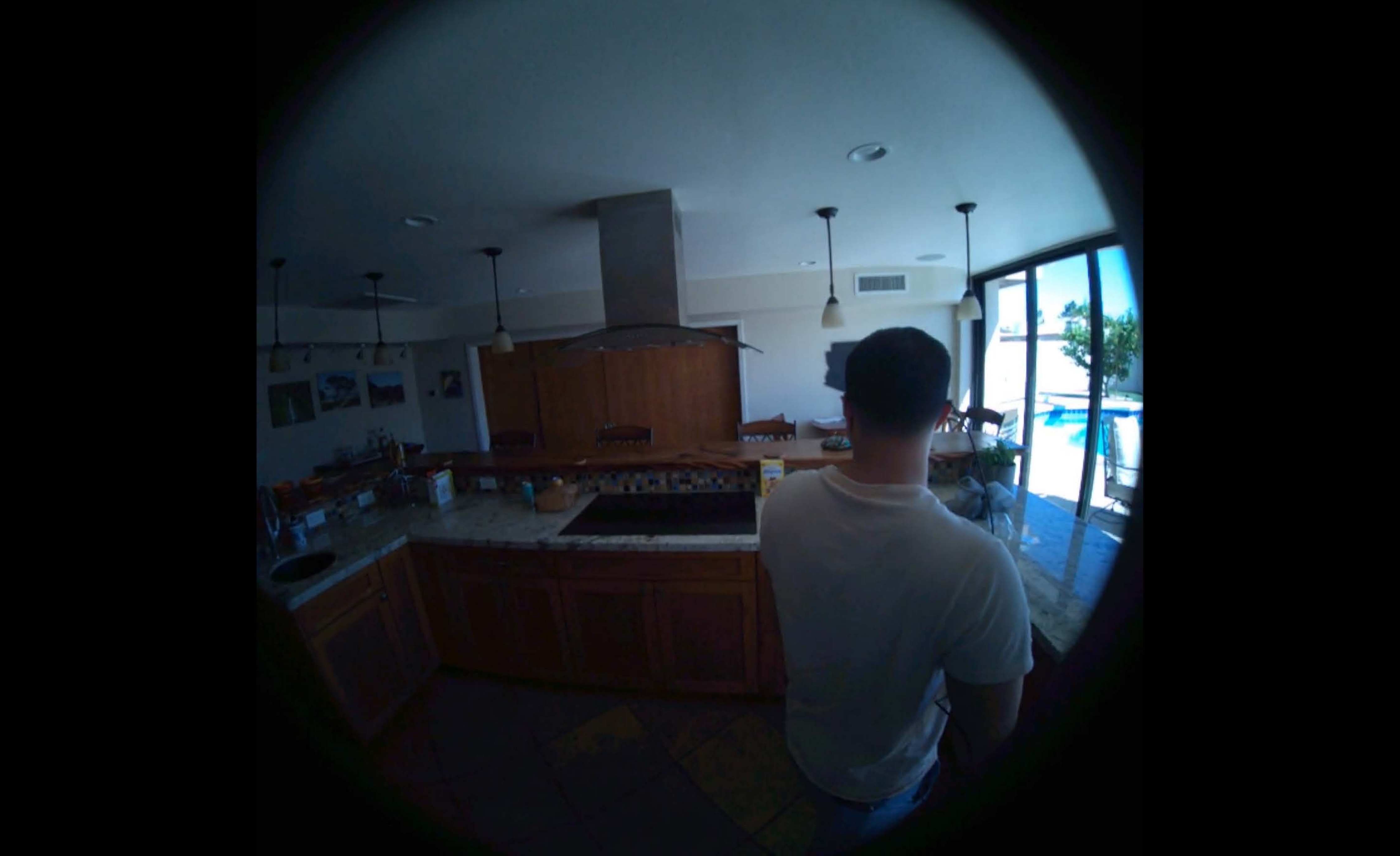}
    \end{minipage}
    \hfill
    \begin{minipage}{0.48\textwidth}
        \centering
        \includegraphics[width=\linewidth]{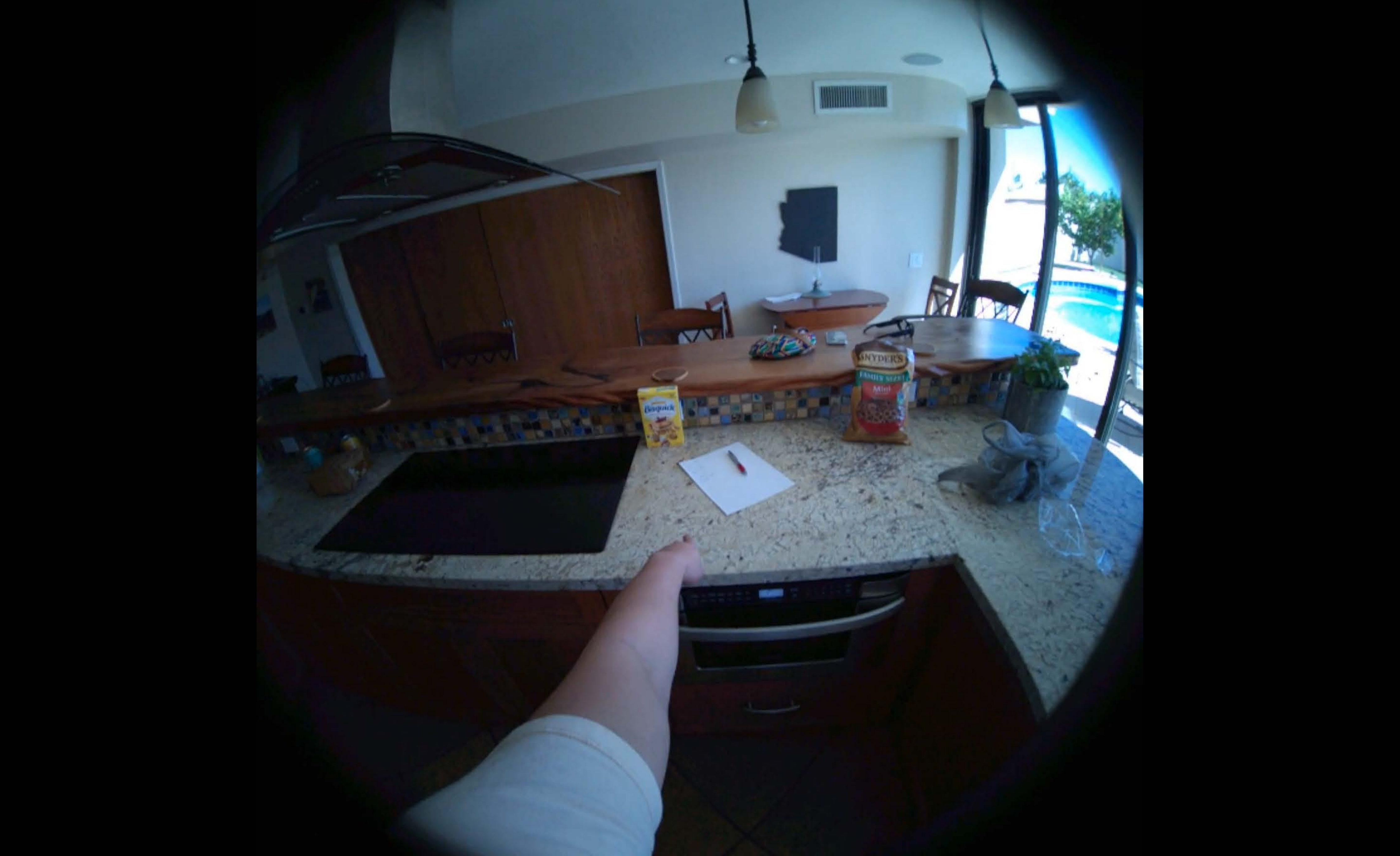}
    \end{minipage}
    \caption{Visual representation of Case 4: Persistent perceptual error leads to an incorrect final reasoning result.}
    \label{fig:case4}
\end{figure}

\textbf{Reasoning Process:}As shown in Figure~\ref{fig:case4}, 1. Between 0:39.975 and 1:14.359, the person interacts with the kitchen island, turns towards the stove, and walks towards the glass doors. 2. Across these consecutive keyframes, the Stage 1 perception layer consistently and incorrectly identifies the target's orientation as \texttt{back-right} instead of the actual \texttt{back-left}. 3. Influenced by this persistent error in visual representation, the Baseline 1 model incorrectly predicts \texttt{back-right}. 4. Anchored by the continuous stream of flawed perceptual data from Stage 1, the Stage 2 reasoning module is unable to break the error chain, ultimately resulting in an incorrect final prediction of \texttt{back-right}.

\end{document}